\documentclass{article}

\usepackage{PRIMEarxiv}

\usepackage[utf8]{inputenc} 
\usepackage[T1]{fontenc}    
\usepackage{hyperref}       
\usepackage{url}            
\usepackage{booktabs}       
\usepackage{amsfonts}       
\usepackage{nicefrac}       
\usepackage{microtype}      
\usepackage{fancyhdr}       
\usepackage{graphicx}       
\graphicspath{{media/}}     

\usepackage{latexsym,bm}
\usepackage{empheq,cases}
\usepackage{floatrow}
\usepackage{tabularx}
\usepackage{multirow}
\usepackage{algorithmic,algorithm}

\title{Sequential Bayesian Design for Efficient Surrogate Construction in the Inversion of Darcy Flows
}

\author{
  Hongji Wang$^a$,\quad Hongqiao Wang$^{*a,b}$, \quad Jinyong Ying$^{a}$
  {and} \quad
  Qingping Zhou$^{a}$ \\
  {\it $^a$School of Mathematics and Statistics} \\
  {\it Central South University} \\
  {\it Changsha 410083, People's Republic of China}\\
  [2mm]
  {\it $^b$Institute of Mathematics} \\
  {\it  Henan Academy of Sciences} \\
  {\it  Zhengzhou 450046, People's Republic of China}\\
  [2mm]
  {$^*$ Corresponding author: Hongqiao Wang}\\
  {\it School of Mathematics and Statistics} \\
  {\it Central South University} \\
  {\it Changsha 410083, People's Republic of China}\\
  {\it E-mail: Hongqiao.Wang@csu.edu.cn}
}

\begin{document}
\maketitle

\begin{abstract}
Inverse problems governed by partial differential equations (PDEs) play a crucial role in various fields, including computational science, image processing, and engineering. 
Particularly, Darcy flow equation is a fundamental equation in fluid mechanics, which plays a crucial role in understanding fluid flow through porous media.
Bayesian methods provide an effective approach for solving PDEs inverse problems, while their numerical implementation requires numerous evaluations of computationally expensive forward solvers.
Therefore, the adoption of surrogate models with lower computational costs is essential. 
However, constructing a globally accurate surrogate model for high-dimensional complex problems demands high model capacity and large amounts of data. 
To address this challenge, this study proposes an efficient locally accurate surrogate that focuses on the high-probability regions of the true likelihood in inverse problems, with relatively low model complexity and few training data requirements.
Additionally, we introduce a sequential Bayesian design strategy to acquire the proposed surrogate since the high-probability region of the likelihood is unknown. 
The strategy treats the posterior evolution process of sequential Bayesian design as a Gaussian process, enabling algorithmic acceleration through \textit{one-step ahead prior}. 
The complete algorithmic framework is referred to as Sequential Bayesian design for locally accurate surrogate (SBD-LAS). 
Finally, three experiments based the Darcy flow equation demonstrate the advantages of the proposed method in terms of both inversion accuracy and computational speed.
\end{abstract}


\section{Introduction}
\label{sec:Introduction}
Partial differential equations (PDEs) play a crucial role in understanding real world phenomena such as complex physical and biological systems. 
What's more, sometimes we need to infer the cause of observations or phenomena that corresponds to the unknown parameters of PDEs, e.g., coefficient functions, initial states, or source terms. 
The process above is called inverse differential problem, which arise in computational science \cite{doi:10.1137/21M1441420,bardsley2014randomize}, image processing \cite{jin2017deep} and engineering domain \cite{kaipio2006statistical}. 
The challenge of inverse differential problem is the substantial computational requirement of the forward solution.

A popular approach nowadays is to use the Bayesian framework to estimate unknown parameters in PDEs. 
This class of methods combines PDEs, discrete observations of functions within the PDEs, and the Bayesian formula to obtain the posterior distribution of unknown parameters in the PDEs. 
From this posterior distribution, estimates of the unknown parameters and their uncertainties can be obtained. 
However, due to the nonlinearity and lack of analytical solutions for most practical PDEs, we are unable to obtain an analytical expression for the posterior distribution. 
As a result, we need to use Markov chain Monte Carlo (MCMC) methods or other other sampling methods to obtain samples from the posterior distribution, which are then used to estimate the unknown parameters of the PDEs.

MCMC methods, particularly those utilizing the Metropolis-Hastings (MH) algorithm \cite{metropolis1953equation,hastings1970monte}, generate Markov chains whose stationary distributions correspond to the Bayesian posterior. 
The option of the MCMC method is simple and the sampling chains will finally converge to the posterior after the long run. 
Another class of methods used in Bayesian inference is variational inference (VI) \cite{blei2017variational}, that a parameterized model is trained to approximate the Bayesian posterior by minimizing the Kullback-Leibler (KL) divergence.
A key challenge of using these methods is finding an optimal trade-off between the quality of the inference and the computational resources required. 
The computational cost primarily arises from numerous calls to the PDE's forward solver.

For systems characterized by large-scale, highly nonlinear parametric PDEs, the computational demand of numerical simulations is significant due to the necessity of repeatedly solving high-dimensional linear systems to achieve a specified level of accuracy. 
Consequently, solving infinite-dimensional Bayesian inverse problems becomes practically infeasible, as approximating these complex infinite-dimensional posterior distributions numerically requires an excessive number of solutions at various parameter settings, suffering from the curse of dimensionality. 
To alleviate the computational challenges associated with these problems, various mathematical and numerical strategies have been developed. 
These include (i) advanced sampling techniques that take advantage of the low-dimensional nature \cite{patel2022solution,cui2016dimension,constantine2016accelerating} or gradients \cite{bui2014solving} of posterior distributions, (ii) methods for direct posterior estimation and statistical analysis, such as Laplace approximation \cite{schillings2020convergence}, deterministic quadrature \cite{schillings2013sparse}, or transport maps \cite{zech2022sparse,wang2022projected}, and (iii) surrogate models using polynomial approximations \cite{amaya2024multifidelity,yan2019adaptive} or reduced-order models \cite{lieberman2010parameter,wang2024total}, often combined with multilevel or multifidelity frameworks \cite{teckentrup2015multilevel,peherstorfer2018survey}.

Neural operators, which are neural network models designed to represent nonlinear maps between function spaces, have recently garnered significant attention. 
These models are highly effective at representing \textit{parameters-to-states} maps defined by nonlinear parametric PDEs and can approximate these maps using only a finite set of solutions at various parameter samples. 
Some of the most prominent neural operators are POD-NN \cite{hesthaven2018non}, DeepONet \cite{lu2019deeponet}, Fourier neural operator \cite{li2020fourier}, and derivative-informed reduced basis neural networks \cite{o2022derivative}. 
The task of approximating nonlinear maps is commonly known as the operator learning problem, while the process of numerically solving this problem through the optimization of neural network weights is referred to as training. 
Thanks to their rapid evaluation speed, neural operators present a promising alternative to traditional surrogate modeling techniques, offering a means to accelerate the posterior inference of infinite-dimensional Bayesian inverse problems. 
This is achieved by replacing the computationally expensive solutions of PDEs with the output of trained neural operators.

Although employing neural network (NN) surrogate models can significantly reduce the computational cost of the posterior inference process, several challenges persist in practical applications. 
These challenges include the selection or construction of an appropriate neural network, the acquisition of training data, and the optimization of the neural network itself. 
To address this issue, some adaptive learning surrogate model approach has emerged \cite{yan2019adaptive,yan2017convergence,gao2024adaptive}. 
These methods leverage an iterative adaptive learning framework, enabling the surrogate model to autonomously identify and update training points, thereby avoiding the need for exhaustive learning over the entire input space. 
This significantly reduces the complexity of training the surrogate model.
In infinite-dimensional Bayesian inverse problems, it is practically impossible to construct a sufficiently accurate surrogate model over the entire domain of the prior distribution due to the vastness of the parameter space. 
This needs high requirements on both the NN and the data. 
Typically, the posterior distribution is concentrated on a small subset of the support of the prior distribution, which is used to infer parameters, while other regions are generally discarded with the probability of almost equal to 0. 
Given this characteristic, this paper primarily focuses on guiding the behavior of NNs by selecting appropriate training data to construct a locally accurate likelihood surrogate. 
Sequential Bayesian, a Bayesian inference method, is widely employed in various studies \cite{papamakarios2016fast, 4276769,chen2023overviewdifferentiableparticlefilters,syae020}, but we haven't seen it used in the inverse problem.
Sequential Bayesian combines Bayesian theory with the concept of sequential data acquisition, where the inference process is not completed in a single step after all the data is obtained, instead, it is continuously updated and refined as data arrives progressively.
In this work, we employ a sequential Bayesian design approach to construct a locally accurate surrogate model, with the aim of reducing the complexity of the surrogate model and the number of training points required in Bayesian inverse problems.

The key contributions and novelty of this work are as follows:
\begin{itemize}
    \item \textbf{Locally accurate surrogate model}: For overcoming the challenge of surrogate in Bayesian inverse problems, demanding high model capacity and large amounts of data, a locally accurate surrogate model is proposed for Bayesian inverse problems, designed to focus on the high-probability regions of the likelihood function. This model provides accurate estimates in the high-probability regions of the true likelihood, while offering rougher estimates in the low-probability areas, contributing to that lower model capacity and less data will suffice.
    \item \textbf{Adaptive experimental design framework SBD-LAS}: Since the true high-probability regions of the likelihood are unknown, we introduce an adaptive experimental design framework that transforms the current posterior information into prior information for the next step. This significantly reduces the surrogate model training time, with more pronounced benefits in complex systems. Additionally, the Gaussian approximation used in the posterior-to-prior transition helps mitigate the effects of multiple peaks in the likelihood, which is common in well-posed problems.
\end{itemize}

The Darcy flow equation is a fundamental equation in fluid mechanics that describes fluid flow through porous media. 
It plays a crucial role in understanding and predicting important processes such as groundwater flow \cite{noorduijn2014representative}, petroleum extraction \cite{siddiqui2016pre}, and the transport of pollutants in environmental engineering \cite{patil2014contaminant}.
Therefore, this paper will adopt Darcy's equation as the experimental system.
More specifically, consider a domain of interest $\Omega=[0,1]^2$, the Darcy's equation is  
\begin{equation}
\label{2dellip}
    \begin{cases}
    -\nabla\cdot(a(\boldsymbol{x}) \nabla u(\boldsymbol{x}))=f(\boldsymbol{x}),\quad \boldsymbol{x}\in \Omega,\\
    \hspace{6em} u(\boldsymbol{x})=0, \hspace{2em}  \boldsymbol{x}\in \partial{\Omega},
    \end{cases}
\end{equation}
where $a$, $u$ and $f$ are the permeability field, the pressure field and the velocity field, respectively.
The data $\boldsymbol{y}$ is given by a finite observation set of $u$, perturbed by noise, and the problem is to recover the permeability field $a$ from these measurements $\boldsymbol{y}$. 
In what follows, we choose the source $f(\boldsymbol{x}) = \sin(\pi x_1)\sin(\pi x_2)$.

In the numerical simulation, we  solve the equation (\ref{2dellip}) using finite element method, in which we can control the accuracy of the solution by adjusting the number of grids. 
The more grids there are, the more accurate the solution is, accompanied by higher computation. 
The computational complexity dramatically increases with the degrees of freedom in grids. 
Therefore, a suitable surrogate model is necessary.

The remainder of the paper is structured as follows.
In Section \ref{sec:Bayesian inversion}, We define the infinite-dimensional Bayesian inverse problem and parameterization of the coefficient function by Gaussian process regression (GPR). 
In Section \ref{sec:NN-based surrogate modeling for PDE solver}, we propose the locally accurate surrogate for inversion problems.
In Section \ref{sec:Adaptive ED for surrogate}, we use Sequential Bayesian design to construct the proposed surrogate. 
Furthermore, we apply Gaussian approximation to priors in Sequential Bayesian design, by which we propose the \textit{one-step ahead prior} to accelerate the Sequential Bayesian design.
In Section \ref{sec:Numerical experiments}, we apply the proposed methods to three numerical experiments, one of which is an interface problem. 
And the other two are inverting a complicated coefficient function and a multi-peak coefficient function, respectively.
Section \ref{sec:Conclusion} provides a brief conclusion and suggests possible future work.

\section{Bayesian inversion}
\label{sec:Bayesian inversion}
\subsection{Infinite dimensional Bayesian inverse problem}
\label{sec:Infinite dimensional Bayesian inverse problem}
In practical applications, infinite-dimensional inverse problems are prevalent across various fields, including signal processing, image reconstruction, and climate modeling. 
These problems typically involve inferring infinite-dimensional parameters or functions from limited observational data. 
These complex problems can all be modeled as a series of systems, which can be precisely described using PDEs or coupled systems of PDEs. 
For simplicity, let $\Omega$ is the domain of systems, $\mathcal{H}(\Omega)$ is an infinite-dimensional separable real Hilbert space, the function $u \in \mathcal{H}(\Omega)$ represents the state of the system, and the function $a \in \mathcal{H}(\Omega)$ denotes the factor affecting system status, called coefficient function. 
Then the system can be described by the following PDE:
\begin{equation}
   \label{Problem}
    \begin{cases}
        &\mathcal{F}(u(\boldsymbol{x});a(\boldsymbol{x})) = 0, \quad \boldsymbol{x}\in \Omega,\\ 
        &\mathcal{B}(u(\boldsymbol{x})) = 0,\quad \boldsymbol{x}\in \partial \Omega,
    \end{cases}
\end{equation}
where $\mathcal{F}$ represents a general partial differential operator, and $\mathcal{B}$ denotes a general boundary operator. 
The solution operator of the PDE is denoted as $\mathcal{A}: \mathcal{H} \rightarrow \mathcal{H}$, the \textit{parameters-to-states} map.
In this context, the inverse problem is inferring the corresponding coefficient function $a$ from the state $u$.

Typically, some discrete and noisy state observation $\boldsymbol{y} \in \mathbb{R}^{d}$ will usually be obtained, which is related to the sate $u$ through an observation operator $\mathcal{O}:\mathcal{H} \rightarrow \mathbb{R}^{d}$ and a noise term $\boldsymbol{\eta}$, within the domain $\Omega$ using some equipment. 
By incorporating the system described in equation \ref{Problem}, the \textit{parameters to observations} map can be defined as
\begin{equation*}
    \label{data_model}
    \boldsymbol{y} = \mathcal{G}(a) + {\boldsymbol{\eta}} := \mathcal{O} \circ \mathcal{A}(a) + {\boldsymbol{\eta}},
\end{equation*}
where $\boldsymbol{\eta} \sim \mathcal{N}(\boldsymbol{0},\Sigma_{\boldsymbol{\eta}})$ is a Gaussian noise with mean vector $\boldsymbol{0}$ and covariance matrix $\Sigma_{\boldsymbol{\eta}}$.

In Bayesian inverse problems, we define the following functional as the likelihood function:
\begin{equation}
    \label{likelihood-function}
    l(a\mid\boldsymbol{y}) = \frac{\exp(-\Phi(a;\boldsymbol{y}))}{Z(\boldsymbol{y})}.
\end{equation}
$\Phi(a;\boldsymbol{y})$ in formula \eqref{likelihood-function} is a important functional for such inverse problems defined as below
\begin{equation}
    \label{least-sqaured-error}
    \Phi(a;\boldsymbol{y}) = \frac{1}{2}\left\|\boldsymbol{y} - \mathcal{G}(a)\right\|^{2}_{\Sigma_{\boldsymbol{\eta}}},
\end{equation}
where $\|\cdot\|_{\Sigma_{\boldsymbol{\eta}}} = \|\Sigma_{\boldsymbol{\eta}}^{-\frac{1}{2}}\cdot\|$ denotes the weighted Euclidean norm in $\mathbb{R}^{d}$. 
And $Z(\boldsymbol{y})$ in formula \eqref{likelihood-function} is the normalization constant defined as
\begin{equation*}
    Z(\boldsymbol{y}) := \int_{\mathcal{H}}\exp(-\Phi(a;\boldsymbol{y}))\mu(\mathrm{d}a).
\end{equation*}

The traditional approach to inverse problems is to obtain the coefficient $a$ corresponding to the observation $\boldsymbol{y}$ by minimizing the formula \eqref{least-sqaured-error}. 
This method typically relies on an initial guess and is sensitive to the choice of initial values, which may lead to local minima rather than the global optimum. 
Furthermore, many inverse problems exhibit high nonlinearity and ill-posedness, complicating the optimization process and increasing computational costs. 
Bayesian inference, on the other hand, offers an alternative approach. 
The Bayesian method treats the parameter $a$ and observation $\boldsymbol{y}$ as two random variables and derives the posterior distribution $\pi_{\text{post}}(a\mid\boldsymbol{y})$ of $a$ given the observation $\boldsymbol{y}$ using Bayes' formula \cite{sprungk2020local} as fellow:
\begin{equation}
    \label{eq:Bayes' theorem}
    \pi_{\text{post}}(a\mid\boldsymbol{y}) \propto l(a\mid\boldsymbol{y}) \cdot \pi_{\text{pri}}(a) \propto \exp(-\Phi(a;\boldsymbol{y})) \cdot \pi_{\text{pri}}(a),
\end{equation}
where $\pi_{\text{pri}}$ is a prior of $a$ on the $\mathcal{H}$. 
In Bayesian inverse problems, the prior is indispensable, as it encapsulates regularity information about the space to which the coefficient function $a$ belongs. 
Typically, the prior distribution is specified in advance. 
In practical applications, Gaussian priors are particularly common, as they define the required regularity by specifying an appropriate covariance function or covariance operator.

The infinite-dimensional Bayesian inverse problem is highly challenging due to the large computational cost associated with solving the forward problem. 
Traditional numerical methods struggle with the exponential growth of computational resources required for posterior computation. 
To address this, surrogate models are crucial, as they provide simplified approximations of complex system behavior, significantly reducing the computational cost. 
These models are particularly beneficial in forward solving, accelerating the Bayesian inference process, numerically implemented by MCMC or VI. 
By integrating surrogate models, the computational burden of infinite-dimensional Bayesian inverse problems can be significantly reduced.

\subsection{Parametric representation of coefficient function by Gaussian process regression}
\label{sec:Parametric representation}
In infinite-dimensional inverse problems, parameterizing the coefficient function to be inverted is a crucial step that significantly affects both the inversion efficiency and stability. 
Common parameterization methods include interpolation-based parameterization \cite{lam1983spatial}, which reduces the dimensionality and is more flexible but may struggle with complex geometric shapes; 
spectral method-based parameterization \cite{grandclement2009spectral}, which efficiently represents smooth functions and converges quickly, but is less suitable for irregular functions; 
and machine learning-based parameterization, which can adapt to complex systems and automatically learn features but requires large amounts of data and may lead to overfitting. 
Each of these methods has its advantages and limitations, and the choice of an appropriate parameterization technique should be made based on the specific problem and the characteristics of the available data.

From a Bayesian standpoint, the coefficient function is regarded as a random variable, following a stochastic process or field. 
In this framework, Gaussian Process Regression (GPR) stands out as an ideal parameterization method, which is a regression technique grounded in random processes and Bayesian principles \cite{williams2006gaussian}, making it a powerful tool for parameterizing random functions in Bayesian inverse problems. 
We parametricize the Gaussian random field directly by function values at fixed discrete locations $X^{*}=$ $\left[\boldsymbol{x}_1^{*}, \ldots, \boldsymbol{x}_p^{*}\right]^{T}$, which we represent using a sparse vector $\boldsymbol{\theta}=\left[a\left(\boldsymbol{x}_1^{*}\right),\ldots,a\left(\boldsymbol{x}_p^{*}\right)\right]^{T}$. 
The parameters of this parameterization that involve location-based covariance, would be advantageous for the preconditioned Crank-Nicolson (pCN) method \cite{cotter2013mcmc}, which is an outstanding MCMC algorithm for image inversion. 
The prior of $a(\boldsymbol{x})$ follows $\mathcal{G} \mathcal{P}\left(0, K^{\boldsymbol{\psi}}(\boldsymbol{x}, \cdot)\right)$, where $\boldsymbol{\psi}$ denote the hyper-parameters in kernel function, and following \cite{williams2006gaussian}, $\forall{\boldsymbol{x}}$, the posterior could be expressed as $\pi(a(\boldsymbol{x}) \mid \boldsymbol{\theta}, \boldsymbol{\psi}) \sim \mathcal{G P}\left(\mu^{\boldsymbol{\psi}}(\boldsymbol{x}), \Sigma^{{\boldsymbol{\psi}}}(\boldsymbol{x})\right)$, where
\begin{equation*}
    \mu^{\boldsymbol{\psi}}(\boldsymbol{x}) = K\left(\boldsymbol{x}, X^{*}\right) K\left(X^{*}, X^{*}\right)^{-1} \boldsymbol{\theta},
\end{equation*}
\begin{equation*}
    \Sigma^{\boldsymbol{\psi}}(\boldsymbol{x}) = K(\boldsymbol{x}, \boldsymbol{x})-K\left(\boldsymbol{x}, X^{*}\right) K\left(X^{*}, X^{*}\right)^{-1} K\left(X^{*}, \boldsymbol{x}\right).
\end{equation*}

Determination of the hyperparameters $\boldsymbol{\psi}$ in kernel function follows the prior $\pi_{\text{pri}}$ of coefficient function $a$ in equation (\ref{eq:Bayes' theorem}). 
Therefore, the inversion of $\pi_{\text {post }}\left(a \mid \boldsymbol{y}\right)$ becomes the inversion of the Gaussian random field governed by $\boldsymbol{\theta}$, expressed as
\begin{equation*}
    \pi_{\text {post}}\left(\boldsymbol{\theta} \mid \boldsymbol{y}\right) \propto l\left(\boldsymbol{\theta} \mid \boldsymbol{y}\right) \pi_{\text {pri}}(\boldsymbol{\theta}),
\end{equation*}
where
\begin{equation}
    \label{likelihood after parameterization}
    l\left(\boldsymbol{\theta} \mid \boldsymbol{y}\right) \propto \exp(-\frac{1}{2}\left\|\boldsymbol{y} - \mathcal{G}(\boldsymbol{\theta})\right\|^{2}_{\Sigma_{\boldsymbol{\eta}}})
\end{equation}
and $\pi_{\text {pri}}(\boldsymbol{\theta})$ is the joint distribution of $\{a(\boldsymbol{x}_{i}^{*})\}_{i=1}^{p}$, and it can be derived from Infinite dimension distribution $\pi_{\text {pri}}(a)$. 
It is important to note that after parameterization, there is a one-to-one correspondence between $\boldsymbol{\theta}$ and $a$. For convenience, we will henceforth refer to $\pi_{\text {pri}}$ as the prior of $\boldsymbol{\theta}$ rather than the prior of $a$, except in special cases. 
And we will introduce the parameterization of piecewise functions for interface problems in section \ref{sec:Interface problem}.

Notably, after parameterizing $a$ with Gaussian Process Regression (GPR), our \textit{parameters to observations} map has been redefined as $\mathcal{G} = \mathcal{O} \circ \mathcal{A} \circ \mathcal{GP}: \mathbb{R}^{p} \rightarrow \mathbb{R}^{d}$, where $\mathcal{GP}: \mathbb{R}^{p} \rightarrow \mathcal{H}$ is a map governed by GPR. 
For clarity, all \textit{parameters to observations} maps mentioned in the following discussion are assumed to incorporate GPR parameterization, such as the PDE solver and the surrogate model, except explicitly stated otherwise.

\section{NN-based surrogate modeling for PDE solver}
\label{sec:NN-based surrogate modeling for PDE solver}
In Bayesian inference, the forward process of solving PDEs is an essential step.
Various numerical methods such as finite difference\cite{kumar2006solution,kushner1976finite,zhao2007compact} and finite element method\cite{szabo2021finite} have been proposed and widely adopted by practitioners for solving PDEs numerically.
However, for more accurate solution of PDEs, the computational complexity of these methods dramatically increases with the degrees of freedom in meshes. 
Solving the linear equations arising from discretizing the PDE with meshes typically requires a computational complexity that scales cubically with the mesh size. 
This growth, scaled to the powers of the spatial dimensions, imposes impractical time requirements for solving high-dimensional PDEs. 
It is unrealistic to directly apply these methods to Bayesian inverse problems, as the Bayesian inference process requires frequent calls to the forward solver.
Therefore, in the Bayesian inverse problem, the surrogate model with low computational cost is usually used instead of the exact PDE solvers, i.e., the finite method.

\subsection{General surrogate modeling}
\label{sec:general surrogate}
In traditional surrogate models, the goal is to provide an accurate estimate of the entire input space. We denote $G: \mathbb{R}^{p} \to \mathbb{R}^{d}$, representing a globally accurate surrogate model, whose training points $\{\boldsymbol{\theta}_{i}\}_{i=1}^{N_{\text{train}}}$ follow 
\begin{equation*}
        \boldsymbol{\theta}_{1}, \boldsymbol{\theta}_{2},..., \boldsymbol{\theta}_{N_{\text{train}}} \stackrel{\text{i.i.d.}}{\sim} U(\Omega).
\end{equation*}
Compared to the exact solver of PDEs, the forward evaluation speed of the surrogate model is significantly faster. 
Once a globally accurate surrogate model is established, it can completely replace the PDE solver in Bayesian inverse problem, providing estimates of the parameterized coefficient function with considerably lower computational cost. 
When using traditional surrogate model, the surrogate likelihood is
\begin{equation}
    \label{surrogate likelihood function}
    \tilde{l}\left(\boldsymbol{\theta} \mid \boldsymbol{y};G\right) \propto \exp(-\frac{1}{2}\left\|\boldsymbol{y} - G(\boldsymbol{\theta})\right\|^{2}_{\Sigma_{\boldsymbol{\eta}}}).
\end{equation}

NNs, as a highly potential and widely studied technology, have found broad applications across various fields. 
In light of this, the present study also adopts neural networks as surrogate models, whose forward solution is very fast. 
However, for data-driven surrogate models, i.e., NNs, constructing a globally accurate model is computationally expensive, especially when dealing with complex problems.
As it requires dense samples in the global space, making the cost comparable to directly using the PDE solver for inference. 

\begin{figure}[H]
    \centering
    \includegraphics[width=1\linewidth]{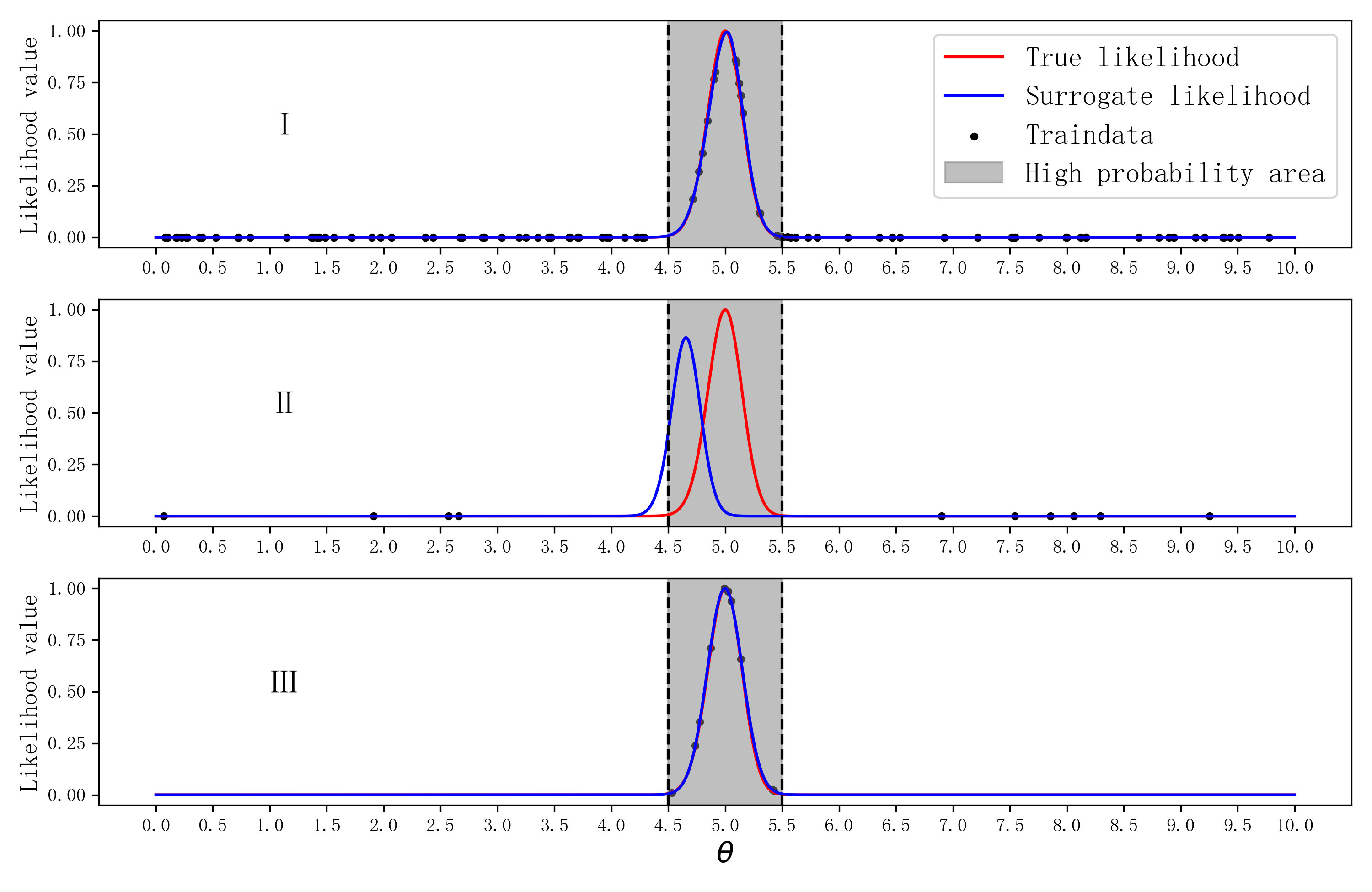}
    \caption{Curves of the likelihood functions under three types of surrogate models: I represents under the globally accurate surrogate model (100 training points, $MSE=0.0003$), II is under the globally coarse surrogate model (10 training points, $MSE=0.3181$), and III is under the locally accurate surrogate model (10 training points, $MSE=0.0002 $). The black dots indicate the locations of the corresponding training points for the surrogate models. The gray shaded areas denote the high-probability regions of the true likelihood.}
    \label{show example}
\end{figure}

Here, we use a one-dimensional ordinary differential equation (ODE) example with a scalar parameter to demonstrate the required number of training points for a globally accurate surrogate model and its corresponding surrogate likelihood performance. 
In this example, the surrogate model, a fully connected neural network, learns the mapping from the \textit{parameters to observations}, and three distinct surrogate models are constructed by using training points sampled from different distributions. 
Further experimental details can be found in \ref{Appendix 1}.
The first picture in Figure \ref{show example} intuitively demonstrates that the globally accurate surrogate needs dense samples in input space. 
To address this issue, in Section \ref{sec:local accurate surrogate}, we propose a specialized surrogate model, locally accurate surrogate, with significantly reduced computational cost, tailored for solving Bayesian inverse problems.

\subsection{Locally accurate surrogate modeling for inverse problem}
\label{sec:local accurate surrogate}
Reviewing the likelihood function $l\left(\boldsymbol{\theta} \mid \boldsymbol{y}\right)$ in equation (\ref{likelihood after parameterization}), it is a probability density function that is higher near the parameter to be inferred and lower elsewhere. 
When a large number of posterior samples are collected, these samples tend to concentrate in these high-probability regions. 
This is because the prior $\pi_{\text{pri}}(\boldsymbol{\theta})$ is typically used for regularization, while the likelihood primarily governs the parameter inference process. 
Therefore, in the context of PDE-based Bayesian inverse problems, the key focus is on the high-probability regions of $l\left(\boldsymbol{\theta} \mid \boldsymbol{y}\right)$, and it suffices to compute the likelihood accurately only in these regions to meet the requirements of the inverse problem.

As discussed above, we propose a locally accurate surrogate model $\tilde{G}: \mathbb{R}^{p} \to \mathbb{R}^{d}$, whose training points $\{\boldsymbol{\theta}_{i}\}_{i=1}^{N_{\text{train}}}$ are sampled from the high-probability regions of $l\left(\boldsymbol{\theta} \mid \boldsymbol{y}\right)$, 
\begin{equation*}
    \boldsymbol{\theta}_{1}, \boldsymbol{\theta}_{2},..., \boldsymbol{\theta}_{N_{\text{train}}} \stackrel{\text{i.i.d.}}{\sim} l\left(\boldsymbol{\theta} \mid \boldsymbol{y}\right).
\end{equation*}
This surrogate model provides accurate estimates in the high-probability regions of $l\left(\boldsymbol{\theta} \mid \boldsymbol{y}\right)$, while offering rough approximations in the low-probability regions. 
When using this locally accurate surrogate model, the corresponding surrogate likelihood function 
\begin{equation*}
    \tilde{l}\left(\boldsymbol{\theta} \mid \boldsymbol{y};\tilde{G}\right) \propto \exp(-\frac{1}{2}\left\|\boldsymbol{y} - \tilde{G}(\boldsymbol{\theta})\right\|^{2}_{\Sigma_{\boldsymbol{\eta}}})
\end{equation*}
matches the true likelihood function $l\left(\boldsymbol{\theta} \mid \boldsymbol{y}\right)$ in the high-probability regions, while still maintaining low probability values in the low-probability regions (as shown picture \uppercase\expandafter{\romannumeral3} in Figure \ref{show example}). 
Consequently, we can use this surrogate likelihood function to approximate the true likelihood function and to invert the parameter of the problem.

It is important to note that the computational cost of the proposed locally accurate surrogate model is significantly lower than that of the traditional globally accurate surrogate model and the former is sometimes even more accurate than the latter in high-probability region of true likelihood. 
Under the same experimental setup as in Section \ref{sec:general surrogate}, Figure \ref{show example} and Table \ref{tab:example_table} clearly demonstrate the performance of the locally accurate surrogate model. 

\begin{table}[H]
    \centering
    \small
    \begin{tabularx}{\textwidth}{p{2cm}>{\centering\arraybackslash}p{1.5cm}>{\centering\arraybackslash}p{1.5cm}>{\centering\arraybackslash}p{1.5cm}>{\centering\arraybackslash}p{1.5cm}>{\centering\arraybackslash}p{1.5cm}>{\centering\arraybackslash}p{1.2cm}}
    \toprule
    \multirow{2}{2cm}{Parameter space} & \multicolumn{2}{c}{Globally accurate} & \multicolumn{2}{c}{Globally rough} & \multicolumn{2}{c}{Locally accurate}\\
    \cline{2-7}
    & $N_{train}$ & MSE & $N_{train}$ & MSE & $N_{train}$ & MSE\\
    \midrule
    $\theta \in (0,10]$ & 100 & 0.0003 & 10 & 0.3181 & 10 & 0.0002\\
    $\theta \in (0,20]$ & 200 & 0.0001 & 10 & 0.2752 & 10 & 0.0002\\
    $\theta \in (0,40]$ & 400 & 0.0064 & 10 & 0.2525 & 10 & 0.0002\\
    \bottomrule
    \end{tabularx}
    \caption{Numbers of the training points and the surrogate likelihoood errors, based on three different surrogate models. $N_{train}$ is the number of the training points. MSE is the index for measuring accuracy of surrogate likelihood in the high-probability region (calculation of MSE can refer to equation (\ref{eq:MSE_of_likelihood}) in \ref{Appendix 1}). First column is the parameter space, which shows the change of $N_{train}$ and MSE as the parameter space increases.}
    \label{tab:example_table}
\end{table}

Specifically, the locally accurate surrogate model requires only 10 training points to achieve $\text{MSE}=0.0002$ in the high-probability region of the true likelihood, where the calculation of $\text{MSE}$ refer to equation (\ref{eq:MSE_of_likelihood}) in \ref{Appendix 1}. 
In contrast, the globally accurate surrogate model, using 100 training points, achieves $\text{MSE}=0.0003$ in the same region. 
Furthermore, as the parameter space expands, especially in the case of high dimensions, the number of training points required by the globally accurate surrogate model increases substantially, whereas the requirements for the locally accurate surrogate model remain unchanged.

\section{Adaptive experimental design for surrogate construction}
\label{sec:Adaptive ED for surrogate}
\subsection{Adaptive Bayesian inference}
\label{sec:Bayesian inference}
In section \ref{sec:local accurate surrogate}, we introduced a locally accurate surrogate model specifically designed for Bayesian inverse problems. 
This model is trained using training points selected from the high-probability regions of the true likelihood function.
However, in practice, these high-probability regions are unknown, making it impossible to directly obtain training points aligned with them. 
To address this challenge, we developed an adaptive experimental design scheme that iteratively updates the prior distribution and the training points drawn from it, facilitating the posterior convergence to the ground truth. 
Ultimately, we are able to obtain samples from the true likelihood and construct the locally accurate surrogate model proposed in section \ref{sec:local accurate surrogate}, which is used in conjunction with the prior $\pi_{\text{pri}}(\boldsymbol{\theta})$ to infer the parameter.

Assuming that, at $(k)$-th iteration, we have the prior $\pi^k_{\text{pri}}(\boldsymbol{\theta})$ and its corresponding surrogate model $G_{k}$ whose training points are samples from $\pi^k_{\text{pri}}(\boldsymbol{\theta})$, the posterior distribution can be inferred using the following formula:
\begin{equation}
\label{eq:pri_post1}
    \pi_{\text{post}}^{k}(\boldsymbol{\theta} \mid \boldsymbol{y}) \propto l^k(\boldsymbol{\theta}) \cdot \pi_{\text{pri}}^{k}(\boldsymbol{\theta}),
\end{equation}
where $l^k(\boldsymbol{\theta}) := \tilde{l}\left(\boldsymbol{\theta} \mid \boldsymbol{y};G_k\right)$ following equation (\ref{surrogate likelihood function}) denotes the $(k)$-th likelihood function. 
A commonly used prior for the next iteration is given by 
\begin{equation}
\label{eq:pri_post2}
    \pi^{k+1}_{\text{pri}}(\boldsymbol{\theta}) = \pi_{\text{post}}^{k}(\boldsymbol{\theta}\mid \boldsymbol{y}),
\end{equation}
indicating that the prior assumption aligns exactly with the posterior distribution from the previous iteration. 
Here, a two-dimensional experiment is conducted to validate the effectiveness of the sequential inference method controlled by Equation (\ref{eq:pri_post1}) and (\ref{eq:pri_post2}). 
For simplicity, the mapping from parameters to observations is defined as a function from two-dimensional space to two-dimensional space, with a NN employed to learn this mapping. 
Specifically, the observation is $\boldsymbol{y}=(11.25, -1.25)$, corresponding to the parameter $\boldsymbol{\theta}^{*}=(2.5, 2.5)$ and  detailed settings are provided in \ref{Appendix 2}. Picture (a) of Figure \ref{fig:show example2-1} illustrates the gradual evolution of the posterior mean under the influence of Equation (\ref{eq:pri_post1}) and (\ref{eq:pri_post2}).
It is worth noting that this inference method is particularly well-suited for the current application scenario. 
This is because the current posterior distribution integrates both prior information and the likelihood, with the prior carrying over information from the previous posterior, including the training points information from the previous step. 
These training points can indicate whether the current location belongs to the high-probability region of the true likelihood. 
Therefore, the surrogate model only needs to learn from the current set of training points at each step, rather than the all training points. 
As is well-known, especially for complex objective functions, surrogate models are more effective at learning in localized regions.

\begin{figure}[H]
    \centering
    \includegraphics[width=1\linewidth]{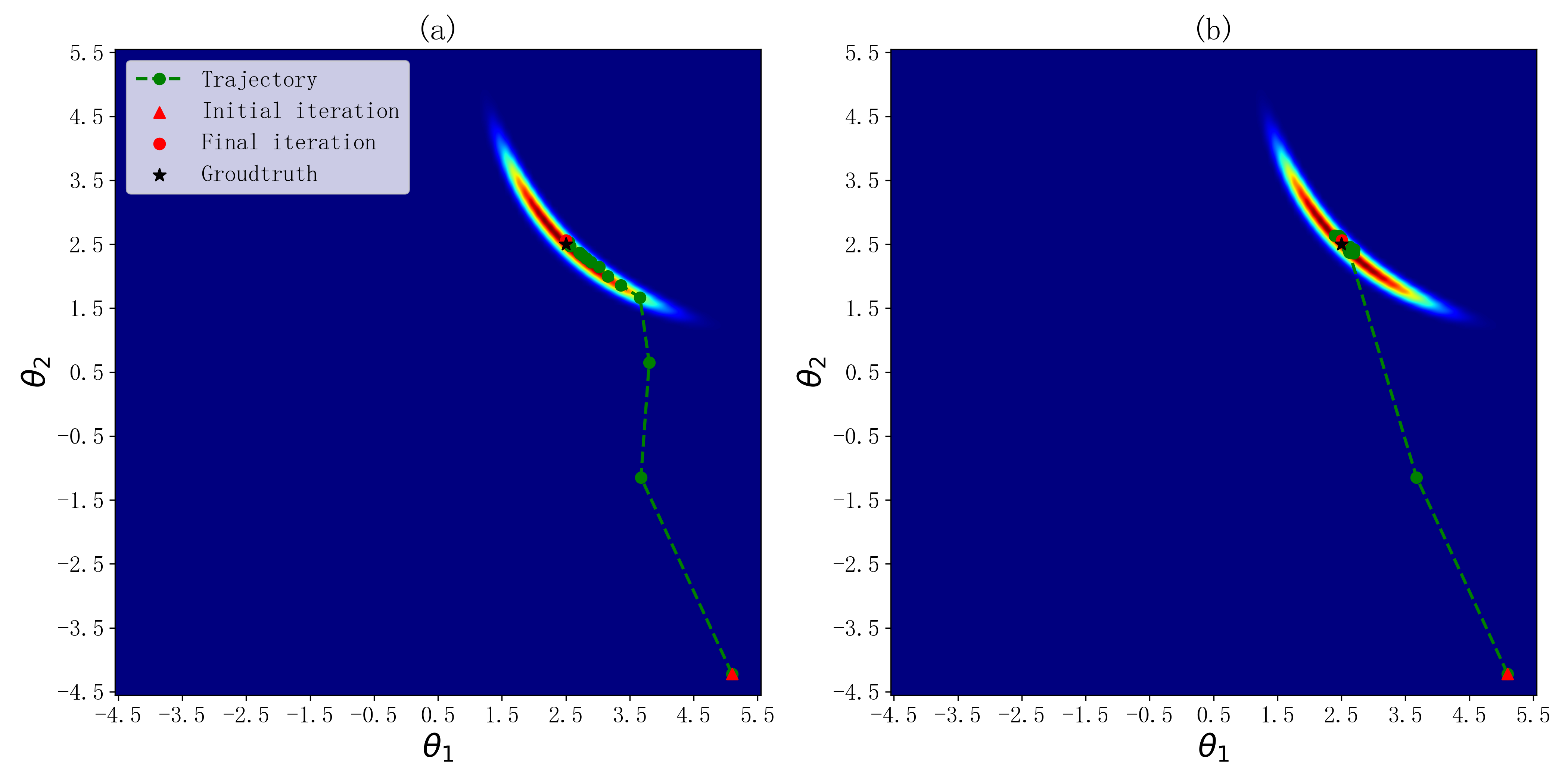}
    \caption{Trajectory of posterior mean points across iterations. The black stars represent the target mean, red triangular points indicate posterior mean points at 1-th iteration, and red circular points mark the final posterior mean points. Green lines and green circular points illustrate the movement trajectory and intermediate points of posterior, respectively. The background shows a heatmap of the target distribution. (a) and (b) are respectively under normal prior and \textit{one-step ahead prior}.}
    \label{fig:show example2-1}
\end{figure}

\subsection{Gaussian approximation and one-step ahead prior}
\label{sec:Design point inference}
Due to the curse of dimensionality, estimating the density of $\pi_{\text{post}}^{\text{k-1}}(\boldsymbol{\theta} \mid \boldsymbol{y})$ becomes infeasible in high-dimensional scenarios.
To overcome this challenge, we propose a simple yet effective approach: utilizing a Gaussian distribution approximation for the prior density at the $(k)$-th iteration based on samples drawn from $\pi^{k-1}_{\text{post}}(\boldsymbol{\theta} \mid \boldsymbol{y})$. 
The Gaussian assumption is particularly well-suited for PDE inversion, as the parameter $\boldsymbol{\theta}$ correspond to mesh points within the spatial domain, naturally inheriting a covariance structure.
Consequently, at the $(k)$-th iteration, the posterior inference can be reformulated as:
\begin{equation}
    \label{eq:update normal pri}
    \pi_{\text{post}}^{k}(\boldsymbol{\theta} \mid \boldsymbol{y}) \propto l^k(\boldsymbol{\theta}) \cdot \mathcal{N}_{\text{pri}}^{k}(\boldsymbol{\theta};\boldsymbol{\mu}_{k},\Sigma_{k}),
\end{equation}
where $\boldsymbol{\mu}_{k}$ and $\Sigma_{k}$ represent the sample mean and sample covariance of $\pi^{k-1}_{\text{post}}(\boldsymbol{\theta} \mid \boldsymbol{y})$, respectively.
Another advantage of the Gaussian approximation is its ability to mitigate the effects of multimodality of $\pi_\mathrm{post}^{k-1}(\boldsymbol{\theta} \mid \boldsymbol{y})$ typically caused by $l^{k-1}(\boldsymbol{\theta})$. 
This simplification facilitates easier sampling of the $(k)$-th posterior $\pi_\mathrm{post}^{k}\left(\boldsymbol{\theta} \mid \boldsymbol{y}\right)$.
Here, we assume that the prior iteration process of the Gaussian approximation $\mathcal{N}_{\text{pri}}(\boldsymbol{\theta};\boldsymbol{\mu},\Sigma)$ is a continuous and gradual correction process, shown visually in (a) of Figure \ref{fig:show example2-1}, meaning that the changes between $\mathcal{N}^{k-1}_{\text{pri}}(\boldsymbol{\theta};\boldsymbol{\mu}_{k-1},\Sigma_{k-1})$ and $\mathcal{N}^{k}_{\text{pri}}(\boldsymbol{\theta};\boldsymbol{\mu}_{k},\Sigma_{k})$ are subtle and incremental.

Here, we propose a \textit{one-step ahead prior} derived from the Gaussian process. 
We assume that the update procedure for $\mathcal{N}_{\text{pri}}^{k}(\boldsymbol{\theta};\boldsymbol{\mu}_{k},\Sigma_{k})$, where $k \in \mathbb{N}_{0}=\{0,1,2,3,\ldots\}$, represents the approximation of the prior of the whole iterations and follows a Gaussian process controlled by $\phi = \{\boldsymbol{\mu}, \Sigma\}$, i.e., $\mathcal{N}_{\text{pri}}(\boldsymbol{\theta}; \phi)$. 
The controlled variable $\phi$ is a function of time $t$, denoted as $\phi(t)$, and the Gaussian approximation of the prior in each iteration is treated as the observation of $\phi$ when $t \in \mathbb{N}_{0}$. 
For each observation of $\phi$, denoted as $\phi_i$, we make a linear assumption, allowing us to compute the prediction of next observation as follows:
\begin{equation}
\label{eq:phi_update}
    \phi_{i+1} \approx \phi_{i} + b, \quad b = \phi_{i} - \phi_{i-1}.
\end{equation}
Furthermore, the strength of the linear assumption can be controlled by adjusting the step ratio, denoted as $\alpha\geq0$, with larger values of $\alpha$ indicating a stronger linear assumption. 
By reformulating equation (\ref{eq:phi_update}), we obtain the following result:
\begin{equation}
\label{eq:phi_update with alpha}
    \phi_{i+1} \approx \phi_{i} + \alpha \cdot b, \quad b = \phi_{i} - \phi_{i-1},
\end{equation}
then we have
\begin{equation}
    \label{eq:update ahead pri}
    \pi_{\text{post}}^{k}(\boldsymbol{\theta} \mid \boldsymbol{y}) \propto l^k(\boldsymbol{\theta}) \cdot \pi_{\text{ahead}}^{k},\quad \pi_{\text{ahead}}^{k} = \mathcal{N}_{\text{pri}}(\boldsymbol{\theta};\phi_{k+1}).
\end{equation}
Notably, the equation (\ref{eq:update ahead pri}) with $\alpha=0$ is equivalent to the equation (\ref{eq:update normal pri}).

We use the same experiment as in Section \ref{sec:Bayesian inference} to demonstrate the effectiveness of the proposed adaptive experimental design framework for inverse problem.
Using the update formula from equation (\ref{eq:update ahead pri}), we performed the same experiment with $\alpha=0$ and $1$, to highlight the acceleration effect of the \textit{one-step ahead prior}.
The Evolution process of the posterior mean and the error of parameter estimate are shown in the Figures \ref{fig:show example2-1} and \ref{fig:show example2-2}, with detailed settings provided in \ref{Appendix 2}.
It can be observed that this adaptive design framework ultimately converges to the true likelihood, and Figure \ref{fig:show example2-2} illustrates the significant acceleration achieved by the \textit{one-step ahead prior}.

\begin{figure}[H]
    \centering
    \includegraphics[width=1\linewidth]{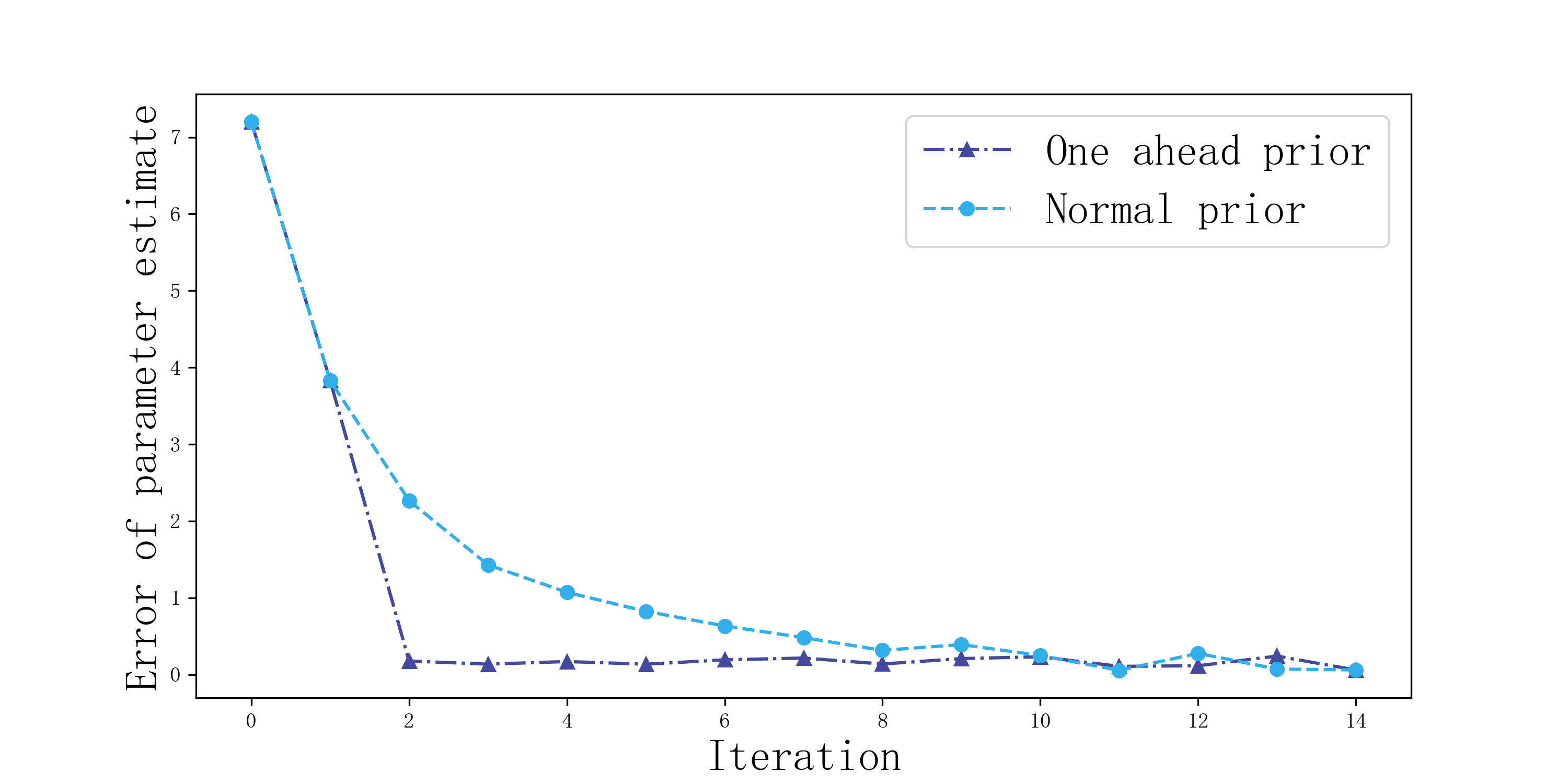}
    \caption{Posterior mean estimation errors vs iterations. Dash-dotted line is based \textit{one-step ahead prior} and dashed line is based normal prior.}
    \label{fig:show example2-2}
\end{figure}

\subsection{The choice of initial prior}
The quality of the initial prior also influences the speed of the experimental point design process. We propose using a coarse PDE solver $\mathcal{G}_{\text{coarse}}$, i.e., finite element method with sparse mesh, to construct the initial prior distribution, as shown below
\begin{equation}
\label{eq:initial pri}
    \pi_{\text{pri}}^{0}(\boldsymbol{\theta}) = \mathcal{N}_{\text{pri}}^{0}(\boldsymbol{\theta};\boldsymbol{\mu}_{0},\Sigma_{0}),
\end{equation}
where ${\boldsymbol{\mu}}_{0} = E[\tilde{l}(\boldsymbol{\theta} \mid \boldsymbol{y}; \mathcal{G}_{\text{coarse}})]$ and $\Sigma_{0} = Cov[\tilde{l}(\boldsymbol{\theta} \mid \boldsymbol{y}; \mathcal{G}_{\text{coarse}})]$ can be estimated using samples of $\tilde{l}(\boldsymbol{\theta} \mid \boldsymbol{y}; \mathcal{G}_{\text{coarse}})$. Although $\mathcal{G}_{\text{coarse}}$ may lack precision, it still retains the fundamental information of the PDE. 
Therefore, using it to construct the initial prior distribution is a reasonable approach. 
By adopting $\pi_{\text{pri}}^{0}(\boldsymbol{\theta})$, we can avoid the randomness associated with blindly starting the adaptive experimental design across the entire space. 
Instead, inference can be started from regions that are closer to the high-probability areas of the true likelihood. 
If we were to start blindly, the initial prior distribution would likely deviate significantly from the true likelihood, necessitating more iterations to converge to the true likelihood, thereby increasing the computational burden.

\subsection{Numerical implementation}
This section provides a detailed description of the complete process for obtaining a locally accurate surrogate model through adaptive experimental design, presented in the form of an algorithm, shown in \textbf{Algorithm \ref{alg:SBD-LAS}}.

\begin{algorithm}[H]
\caption{Sequential Bayesian design for locally accurate surrogate (SBD-LAS)}
\label{alg:SBD-LAS}
\begin{algorithmic}[1]
    \STATE {\textbf{Algorithm Parameters}: fine system solver $\mathcal{G}$, coarse system solver $\mathcal{G}_{\text{coarse}}$, state observation $\boldsymbol{y}$, prior $\pi_{\text{pri}}(\boldsymbol{\theta})$, number of training points per iteration $M$, total iterations of algorthm $K$, step ratio $\alpha$;} 
    \STATE {\textbf{Output}: local accurate surrogate model $\tilde{G}$, estimate of parameter $\hat{\boldsymbol{\theta}}$;}  
    \STATE Draw samples $\Theta$ from $\tilde{l}(\boldsymbol{\theta} \mid \boldsymbol{y}; \mathcal{G}_{\mathrm{coarse}})$; 
    \STATE Get $\pi_{\text{pri}}^0$ following equation (\ref{eq:initial pri}); \hfill (\textit{Initialize prior})
    \STATE Draw $M$ training points $D=\{\boldsymbol{\theta}_i,\mathcal{G}(\boldsymbol{\theta}_i)\}_{i=1,...,M}$ from $\pi_{\text{pri}}^0$;  \hfill (\textit{Initialize training points})
    \STATE Train the surrogate model $G_{0}$ using $D$ and obtain the $l^{0}\left(\boldsymbol{\theta}\right)$; \hfill (\textit{Initialize surrogate})
    \FOR {$k=0$ to $K-1$}
        \STATE Let $\pi_{\text{post}}^{k}(\boldsymbol{\theta} \mid \boldsymbol{y}) = l^{k}\left(\boldsymbol{\theta}\right) \cdot \pi_{\text{pri}}^{k}$; \hfill (\textit{Update posterior})
        \STATE Draw samples $\Theta^{k}$ from $\pi_{\text{post}}^{k}(\boldsymbol{\theta} \mid \boldsymbol{y})$;
        \STATE Compute $\phi_{k}$ by $\Theta^{k}$; 
        \STATE Compute prediction $\phi_{k+1}$ using the formula (\ref{eq:phi_update with alpha}) and get the $\pi_{\text{ahead}}^{k}$;
        \STATE Let $\pi_{\text{pri}}^{k+1} = \pi_{\text{ahead}}^{k}$; \hfill (\textit{Update prior})
        \STATE Draw $M$ training points $D=\{\boldsymbol{\theta}_i,\mathcal{G}(\boldsymbol{\theta}_i)\}_{i=1,...,M}$ from $\pi_{\text{pri}}^{k+1}$; \hfill (\textit{Update training points})
        \STATE Train the surrogate model $G_{k+1}$ using $D$ and obtain the $l^{k+1}\left(\boldsymbol{\theta}\right)$; \hfill (\textit{Update surrogate})
    \ENDFOR
    \STATE Draw samples $\Theta^{K}$ from $l^{K}(\boldsymbol{\theta})\cdot\pi_{\text{pri}}(\boldsymbol{\theta})$; \hfill (\textit{Invert parameter})
    \STATE Let $\tilde{G} = G_{K}$, $\hat{\boldsymbol{\theta}}$ is the mean of samples $\Theta^{K}$;
    \STATE \textbf{Return}: $\tilde{G}$, $\hat{\boldsymbol{\theta}}$;    
\end{algorithmic}
\end{algorithm}

After obtaining a reasonably $\pi_{\text{pri}}^{0}$ using a computationally inexpensive coarse solver, based on the update formula (\ref{eq:update ahead pri}) in Section \ref{sec:Design point inference}, we can iteratively update the surrogate model and the prior. 
When it converges, we can collect sample points from the high-probability region of the true likelihood function and use these samples to train the locally accurate surrogate model introduced in Section \ref{sec:local accurate surrogate}. 
Subsequently, this locally accurate surrogate model, along with the prior distribution $\pi_{\text{pri}}$ of the inverse problem, can be used to efficiently infer the unknown parameters.

It is important to emphasize that, during the process of obtaining posterior samples at the ($k$)-th interation, different sampling algorithms can be chosen, such as variational inference, MCMC methods, rejection sampling, among others.
In this study, we employ the preconditioned Crank-Nicolson (pCN) \cite{cotter2013mcmc} sampling algorithm from the MCMC family, which is particularly well-suited for high-dimensional Bayesian inference problems, as its proposal distribution directly employs the structural information of prior.

\textit{Remarks}:

\begin{itemize}
    \item The linear assumption is quite stringent for the dynamic variation of the Gaussian prior covariance. Therefore, in practical applications, we perform linear prediction only on the prior mean, while the prior covariance is not predicted. Instead, we directly use the covariance value observed at the current step.
    \item To enhance sampling efficiency, each time the pCN algorithm is used to collect posterior samples, the mean of the current prior distribution can be chosen as the starting point for the random walk process.
    \item For the neural network surrogate model, the surrogate model $G_{k+1}$ at the $(k+1)$-th step can be trained using new training points, building upon the surrogate model $G_{k}$ at the $(k)$-th step. This process is akin to adjusting the previous surrogate model with the new training points, significantly accelerating the training process of the surrogate model.
\end{itemize}

\section{Numerical experiments}
\label{sec:Numerical experiments}
In this section, we demonstrate the proposed adaptive experimental design for locally accurate surrogate (SBD-LAS) on the Darcy flow equation, described in more detail at section \ref{sec:Introduction}.
Notably, the complexity of inverse problems is largely influenced by the geometric complexity of the coefficient function, particularly in scenarios involving interfaces. 
To systematically create inverse problems of varying difficulty, we designed three coefficient functions with different shapes and complexities as inversion targets, and generated corresponding observational data using a high-precision PDE solver (the finite method with the dense mesh), with the last one being an interface problem. 
The surrogate model used in the following experiments combines a coarse solver (the finite method with the sparse mesh) with a residual network, where the residual network is employed to correct the bias introduced by the coarse solver, with detailed implementation available in the reference \cite{hu2024mcmc}.
Importantly, our approach imposes no specific restrictions on the choice of surrogate model, as long as it is data-driven, such as Fourier Neural Operators (FNO) \cite{li2020fourier} and Deep Operator Networks (DeepONet) \cite{lu2019deeponet}.

In this study, we predefine the prior distribution of the coefficient function, assuming it follows a Log-Gaussian random field, denoted as
\begin{equation}
\label{eq:log-gaussian random field}
    ln(a(\boldsymbol{x})) \sim \mathcal{N}(\mu(\boldsymbol{x}), k(\boldsymbol{x}^{'},\cdot)), \quad \mu(\boldsymbol{x}) = 0, \quad k(\boldsymbol{x}^{'},\boldsymbol{x}) = \gamma \cdot \text{exp}\left({-\frac{\|\boldsymbol{x}^{'}-\boldsymbol{x}\|_2}{2l^2}}\right).
\end{equation}
The same prior settings are used for all three numerical experiments. 
In the experiments \ref{sec:the first experiment} and \ref{sec:second experiment}, we let $\gamma=1$ and $l=0.5$, and the prior $\pi_{\text{pri}}(\boldsymbol{\theta})$ can be derived. 
In the experiments \ref{sec:Interface problem}, we let $\gamma=1$ and $l=1$, and the prior $\pi_{\text{pri}}(\boldsymbol{\theta})$ can also be derived. 
In practical applications, different forms of kernel functions and their hyperparameters can be selected based on the specific regularity requirements.

We will next demonstrate the inference capacity of the proposed method, the acceleration effect of the \textit{one-step ahead prior}, and the robustness of the method to noise through a series of experiments.
In this study, the posterior mean is used to estimating the unknown coefficient function.
Notably, all coefficient function images in the same experiment have uniform colorbar, which is shown beside the image of true coefficient function. 
The estimation error is defined as follows 
\begin{equation}
\label{eq:MSE}
    error = \frac{\| \hat{\boldsymbol{\theta}} - \boldsymbol{\theta}^{*} \|_{2}^{2}}{p},
\end{equation}
where $\boldsymbol{\theta}^{*}$ and $\hat{\boldsymbol{\theta}}$ are the true parameter and estimated parameter, respectively, and $p$ is the dimension of $\boldsymbol{\theta}$. Moreover, we will use the mean of the posterior sample as the parameter estimation in experiments.

It is worth noting that, during the prior update process, we can add a diagonal matrix $I \cdot \text{c}^{2}$ (c is a constant) to the prior covariance matrix to regulate the diversity of training points generated by the prior distribution, as well as the weight of the prior in the posterior update. 
This technique aids in fine-tuning the prior distribution, allowing it to more accurately reflect its intended influence in the sequential Bayesian inference process.

All experiments in this study were implemented using the Python programming language. The construction, training, and prediction of the NN were carried out with the PyTorch library. For the finite element method (FEM) solution of PDEs, the specialized FEniCS library was employed. In addition, due to resource constraints, only CPU was used in our experiments.

\subsection{A complicated coefficient function}
\label{sec:the first experiment}
\begin{figure}[H]
    \centering
    \includegraphics[width=1\linewidth]{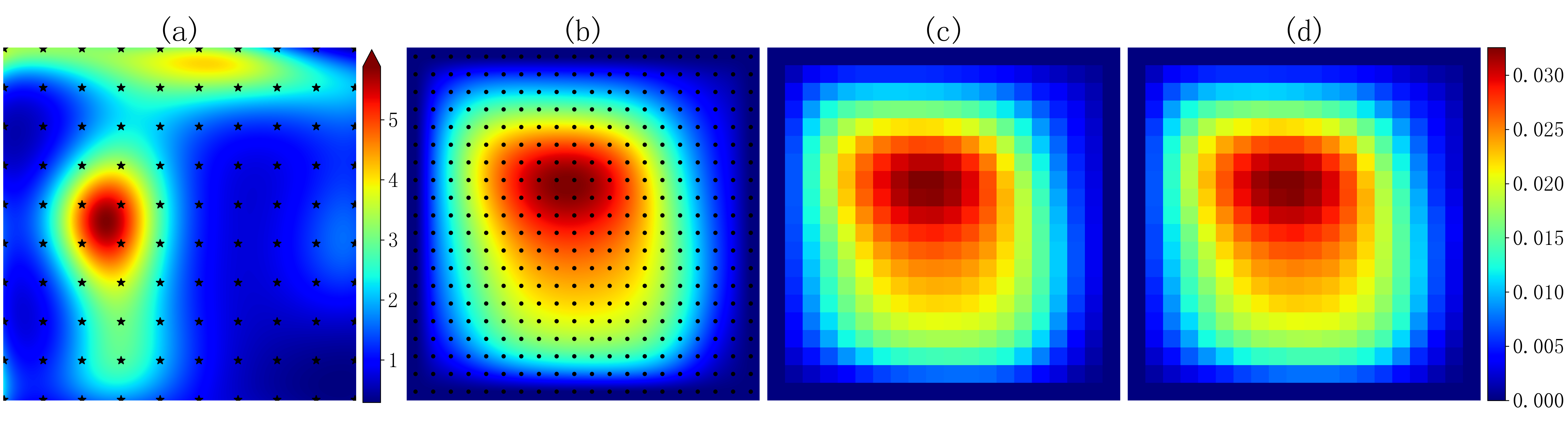}  
    \caption{The true images of the first experiment: (a) is true coefficient $a$ with black stars representing positions of parameter $\boldsymbol{\theta}$, (b) is corresponding solution $u$ with black dots representing locations of observation, (c) and (d) are noise-free observation $\boldsymbol{y}$ and noisy observation $\boldsymbol{y}^{*}$ with noise standard deviation $\delta=0.0001$, respectively.}
    \label{fig:Experment1_ture_a_u}
\end{figure}

Firstly, we present our first experiment.  
The image (a) in Figure \ref{fig:Experment1_ture_a_u} is the true coefficient function whose parameter $\boldsymbol{\theta}$ is a random sample of prior $\pi_{\text{pri}}(\boldsymbol{\theta})$, in the meanwhile, $\boldsymbol{\theta}$ is also the values of $a$ at $10\times10$ grid nodes within $\Omega$. 
The corresponding solution to the PDE is presented in Figure \ref{fig:Experment1_ture_a_u} (image (b)). 
We measure the $u(x)$ at equispaced grid nodes within $\Omega$, to obtain noise-free observation $\boldsymbol{y}$ (image (c) in Figure \ref{fig:Experment1_ture_a_u}), and then introduce observational noise (noise standard deviation $\delta=0.0001$) to obtain noisy observation $\boldsymbol{y}^*$ (image (d) in Figure \ref{fig:Experment1_ture_a_u}). 
We perform inversions on both $\boldsymbol{y}$ and $\boldsymbol{y}^*$, using proposed method SBD-LAS.

Here, we outline the various settings of the method. 
The residual network used has 3 hidden layers with 500 units each and the activation function uses Sigmoid. 
The coarse and fine solvers are finite element methods on $7\times7$ and $20\times20$ grids, respectively. 
The total iterations of SBD-LAS is set to $K = 10$, with $N = 50000$ pCN steps and a step size of $\beta = 0.008$, and the number of training points is $M = 500$ for each iteration.
The initial prior $\pi_{\text{pri}}^{0}(\boldsymbol{\theta})$ is obtained by running  200,000 steps pCN with coarse solver. 

\begin{figure}[H]
    \centering
    \includegraphics[width=0.25\linewidth]{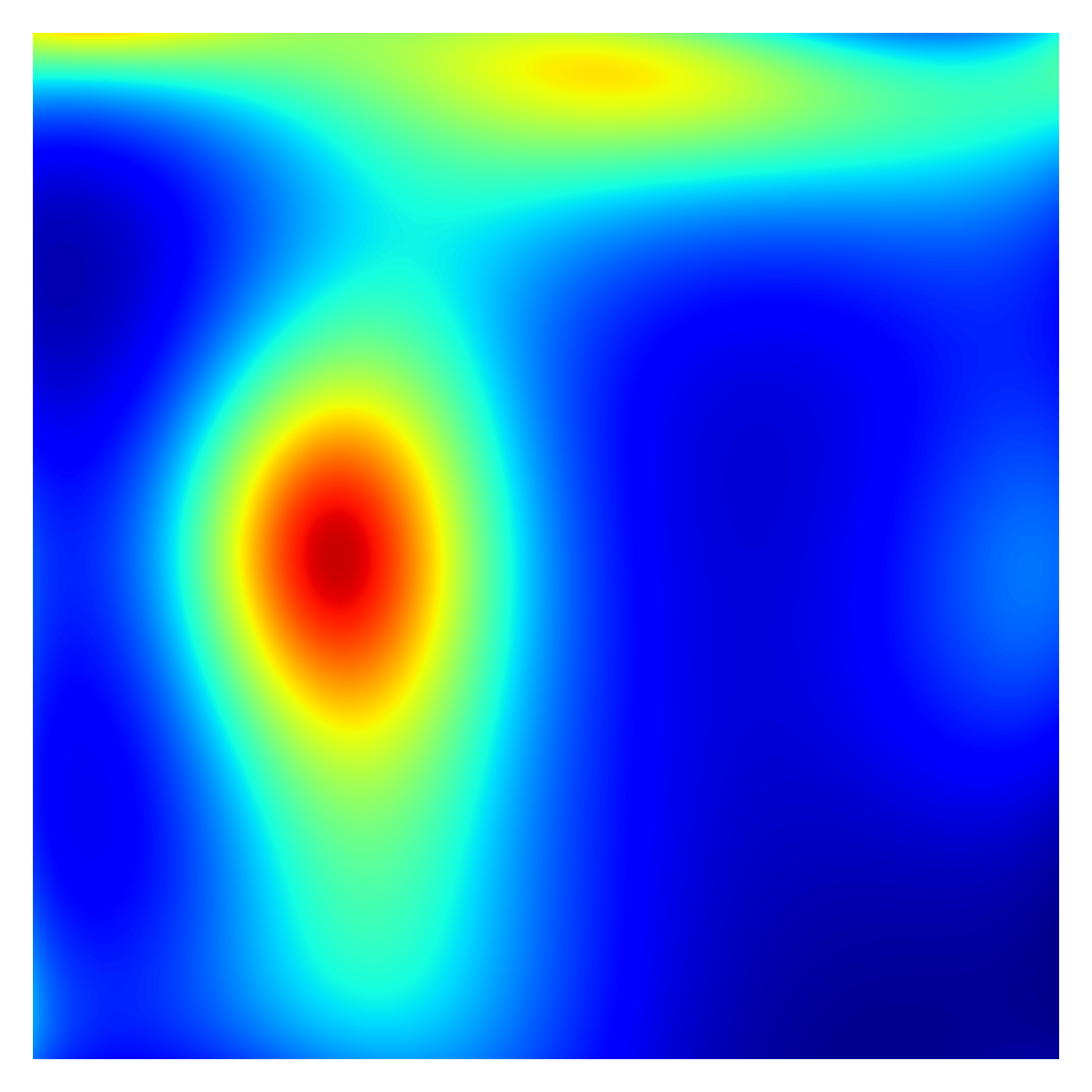}
    \caption{(Experiment 1) Inverse image of noise-free $\boldsymbol{y}$, based on pCN with the fine solver.}
    \label{fig:Experment1_estimate_fine}
\end{figure}

\begin{figure}[H]
    \centering
    \includegraphics[width=1\linewidth]{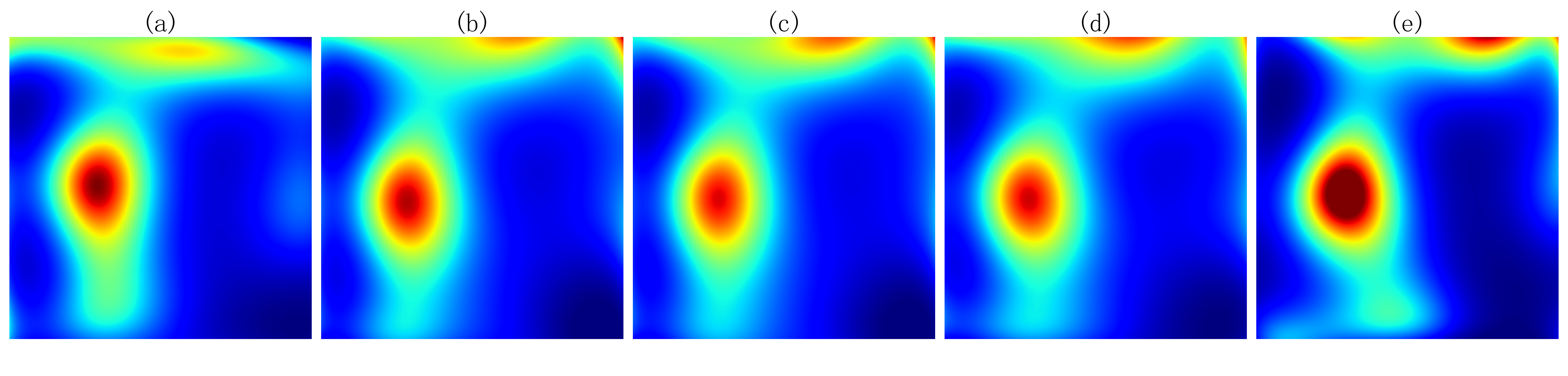}
    \caption{(Experiment 1) Inverse images of noise-free $\boldsymbol{y}$ by SBD-LAS. Images (a) and (e) are true coefficient function and inversion result based on pCN with the coarse solver,  respectively. Images (b), (c) and (d) are inversion results obtained by SBD-LAS with step ratios $\alpha=0, 0.1, 0.5$, respectively.}
    \label{fig:Experment1_estimate_SBD-LAS}
\end{figure}

Figure \ref{fig:Experment1_estimate_fine} shows the estimated coefficient function obtained by running pCN with the fine solver, with samples drawn from 500,000 pCN steps, which serve as the benchmark for evaluating the performance of the proposed method SBD-LAS.
Figure \ref{fig:Experment1_estimate_SBD-LAS} presents the coefficient function estimated by the SBD-LAS, with step ratios $\alpha=0, 0.1, 0.5$.
As seen, the proposed method accurately captures the shape of the coefficient function, including the peaks and valleys, and provides more precise estimates compared to the coarse solver.
The final error values by SBD-LAS with $\alpha=0, 0.1, 0.5$ are 0.2597, 0.2471, and 0.2354, respectively, all significantly smaller than the error of estimation by using pCN with the coarse solver (0.4609), and close to the error of using pCN with fine solver (0.1976), more detailed compare is shown in the Table \ref{tab:Experiment1_MSE_FineSolverNums}.
The SBD-LAS required 5,000 calls to the fine solver, a substantial reduction compared to the 500,000 calls needed using pCN with the fine solver, demonstrating the effectiveness of the proposed method.
When the computational cost of the fine solver is high, the efficiency gains are particularly notable.
While the coarse solver can generally capture the shape of the coefficient function, it fails to identify the peaks, whereas our method successfully find both the peaks and valleys.
The computation time of using pCN with the fine solver is 6470s, and the counterparts of the SBD-LAS with $\alpha=0$, $\alpha=0.1$, $\alpha=0.5$ are 3589s, 3828s, 4245s. 
The computation time of using pCN with the fine solver includes the time of the fine solver call and the time of GPR parameterization. 
The computation time of the SBD-LAS includes the time of the surrogate call, GPR parameterization, training surrogate, getting training data and getting the $\pi_{\text{pri}}^{0}(\boldsymbol{\theta})$.

\begin{table}[H]
    \centering
    \small
    \begin{tabularx}{\textwidth}{p{1cm}>{\centering\arraybackslash}p{2cm}>{\centering\arraybackslash}p{2cm}>{\centering\arraybackslash}p{2cm}>{\centering\arraybackslash}p{2cm}>{\centering\arraybackslash}p{2cm}}
    \toprule
     & \textbf{Fine solver} & \textbf{Coarse solver} & \textbf{SBD-LAS $\alpha=0$} & \textbf{SBD-LAS $\alpha=0.1$} & \textbf{SBD-LAS $\alpha=0.5$} \\
    \midrule
    $\textbf{error}$ & 0.1976 & 0.4609 & 0.2597 & 0.2471 & 0.2354 \\
    $\boldsymbol{N_f}$ & 5e5 & 0 & 5e3 & 5e3 & 5e3 \\
    \bottomrule
    \end{tabularx}
    \caption{(Experiment 1) The errors of parameter estimation and the numbers of calls to the fine solver $N_f$, based on different methods.}
    \label{tab:Experiment1_MSE_FineSolverNums}
\end{table}

\begin{figure}[H]
    \centering
    \includegraphics[width=0.8\linewidth]{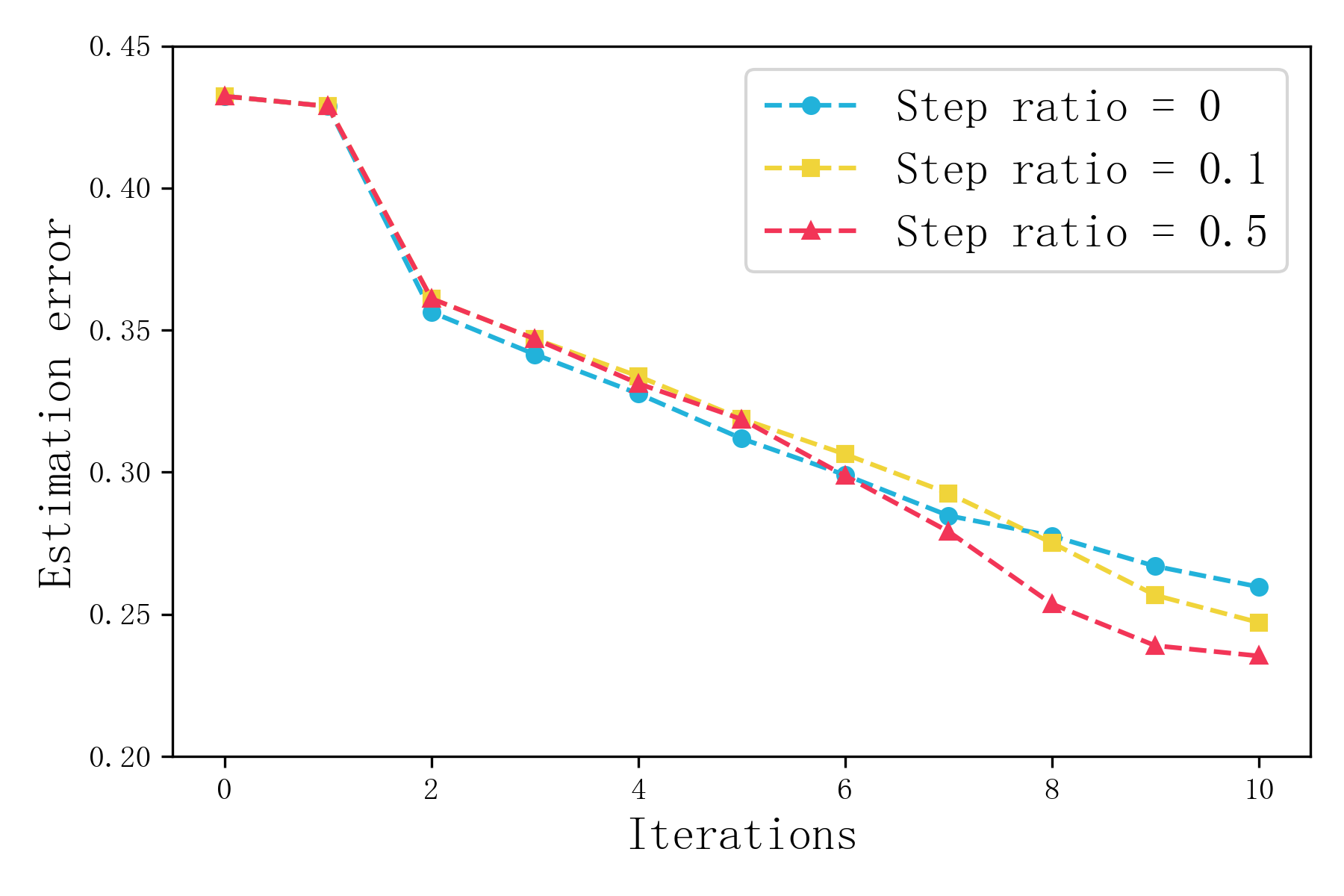}
    \caption{(Experiment 1) Errors of parameter estimation vs iterations. The three curves are from proposed method SBD-LAS with three different step ratios $\alpha=0,0.1,0.5$. The calculation of error follows formula (\ref{eq:MSE}).}
    \label{fig:Experment1_MSE_vs_iter}
\end{figure}

Figure \ref{fig:Experment1_MSE_vs_iter} illustrates the trend of errors of parameter estimates throughout the iterative process, highlighting the significant acceleration effect of the \textit{one-step ahead prior}. 
At iteration 0, the error corresponds to the estimate obtained using the coarse solver. Since at least two estimates are required to implement the \textit{one-step ahead prior}, the error remains the same for the first two iterations regardless of the step ratio $\alpha$. 
It is clearly evident from the Figure \ref{fig:Experment1_MSE_vs_iter} that the \textit{one-step ahead prior} not only accelerates the convergence rate but also achieves higher accuracy after convergence compared to the case without the \textit{one-step ahead prior}. 
In addition, by SBD-LAS method, only 8 iterations in case $\alpha=0.5$ can achieve the same accuracy as 10 iterations in case $\alpha=0$, and only 9 iterations in case $\alpha=0.1$ can achieve the same accuracy as 10 iterations in case $\alpha=0$.

\begin{figure}[H]
    \centering
    \includegraphics[width=0.6\linewidth]{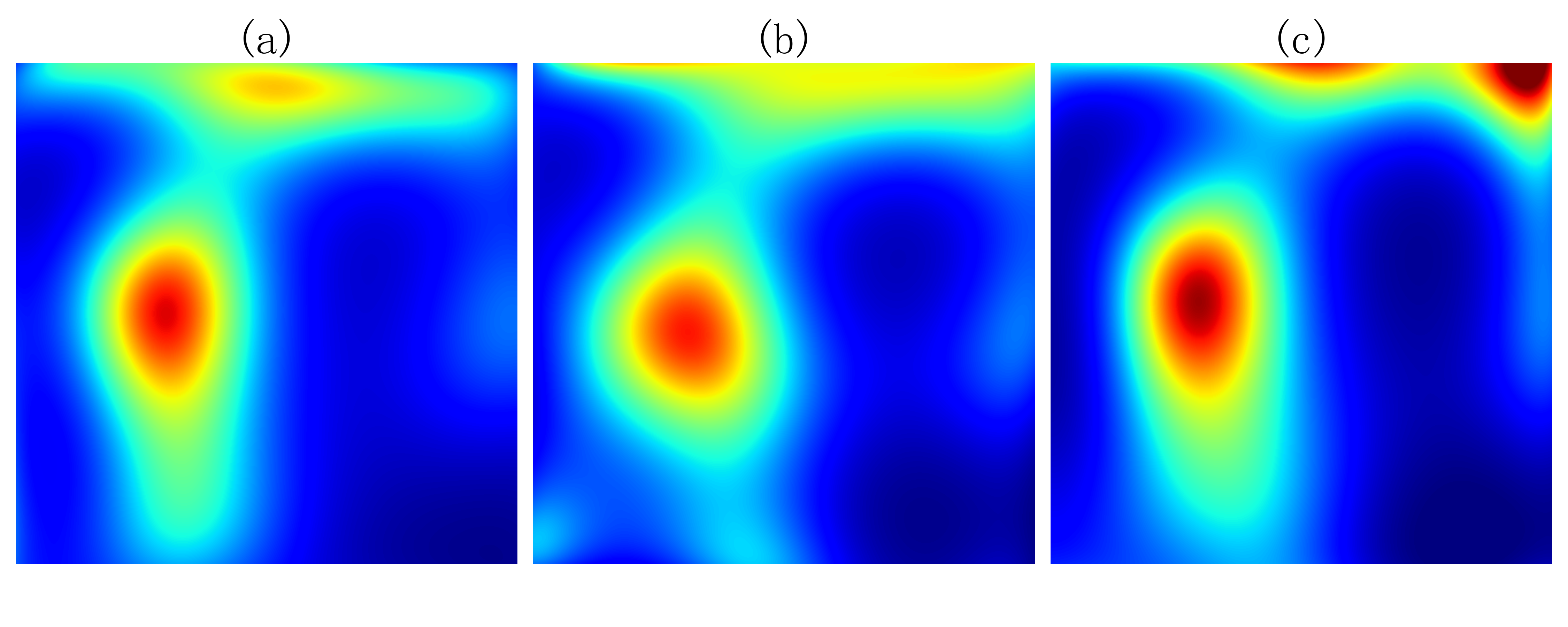}
    \caption{(Experiment 1) Inverse images of noisy observation $\boldsymbol{y}^{*}$ ($\delta=0.0001$). Images (a), (b), (c) are inversion results obtained by pCN with the fine solver, SBD-LAS with step ratio $\alpha=0$ and pCN with the coarse solver, respectively.}
    \label{fig:Experment1_estimates_noisy}
\end{figure}

Finally, we applied the proposed SBD-LAS method to invert noisy observational data, with a noise level $\delta=0.0001$ and a step ratio $\alpha=0$ for SBD-LAS. 
The iterative process was carried out for iteration number $K=10$, with 1000 training points in each iteration, and the settings for the surrogate model were consistent with those used in the noise-free experiment. 
The estimated coefficient functions are shown in the Figure \ref{fig:Experment1_estimates_noisy} and error of the estimates for the fine solver, SBD-LAS, and the coarse solver were 0.2106, 0.2779, and 0.3639,  respectively. 
These results demonstrate that our method remains effective in handling noisy observation.

\subsection{A multi-peak coefficient function}
\label{sec:second experiment}
\begin{figure}[H]
    \centering
    \includegraphics[width=1\linewidth]{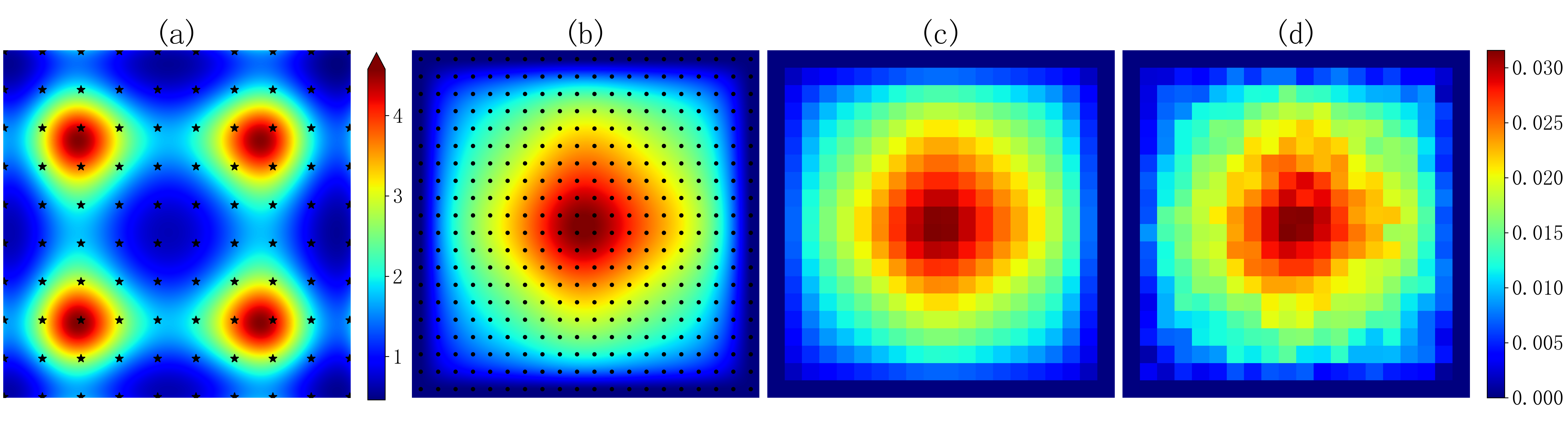}
    \caption{The true images of the second experiment: (a) is true coefficient $a$ with black stars representing positions of parameter $\boldsymbol{\theta}$, (b) is corresponding solution $u$ with black dots representing locations of observation, (c) and (d) are noise-free observation $\boldsymbol{y}$ and noisy observation $\boldsymbol{y}^{*}$ with noise standard deviation $\delta=0.001$, respectively.}
    \label{fig:Experment2_ture_a_u}
\end{figure}

In this experiment, the goal is to assess the proposed method's ability to capture the peaks and valleys of the coefficient function. 
The prior setup for the coefficient function is identical to that in Experiment \ref{sec:the first experiment}, following equation \ref{eq:log-gaussian random field}. 
However, to introduce more peaks and valleys, the parameter $\boldsymbol{\theta}$ of the true coefficient function are not randomly selected from the prior distribution $\pi_{\text{pri}}(\boldsymbol{\theta})$, instead, directly derived from the values of the function $h(x_1,x_2) = 0.5sin(4\pi(x_1-0.1)) + 0.5sin(4\pi(x_2-0.1)) + 0.5$ at the parameterized locations. 
The true coefficient function, its corresponding PDE solution, noise-free observation $\boldsymbol{y}$ and noisy observation $\boldsymbol{y}^{*}$ (noise standard deviation $\delta=0.001$) are shown in Figure \ref{fig:Experment2_ture_a_u}. 
The configurations for the surrogate model, fine solver, and coarse solver remain the same as in Experiment \ref{sec:the first experiment}.
For the SBD-LAS method, a slight adjustment was made to the setup from Experiment \ref{sec:the first experiment}, with the iteration number $K$ set to 21 and he number of training points $M$ set to 1000 for each iteration, while the other settings remained unchanged.
And the initial prior $\pi_{\text{pri}}^{0}(\boldsymbol{\theta})$ is obtained by running  500,000 steps pCN with coarse solver. 

\begin{figure}[H]
    \centering
    \includegraphics[width=0.25\linewidth]{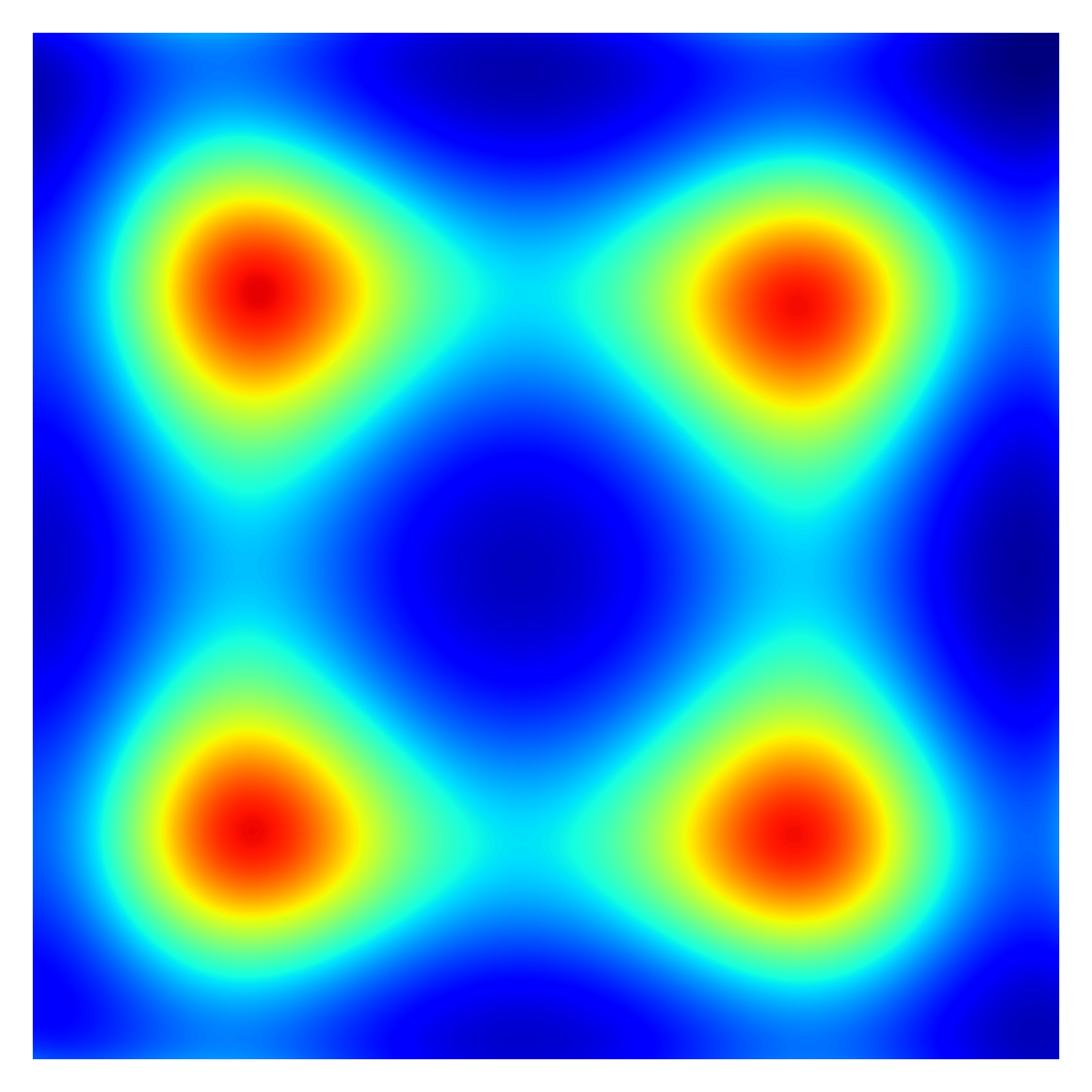}
    \caption{(Experiment 2) Inverse image of noise-free $\boldsymbol{y}$, based on pCN with the fine solver.}
    \label{fig:Experment2_estimate_fine}
\end{figure}

Figure \ref{fig:Experment2_estimate_fine} presents the inversion results obtained by running the pCN algorithm with the fine solver, where the pCN algorithm was executed for 500,000 steps. 
Similarly, the results obtained by pCN with the fine solver are used as the benchmark for comparison.

\begin{figure}[H]
    \centering
    \includegraphics[width=1\linewidth]{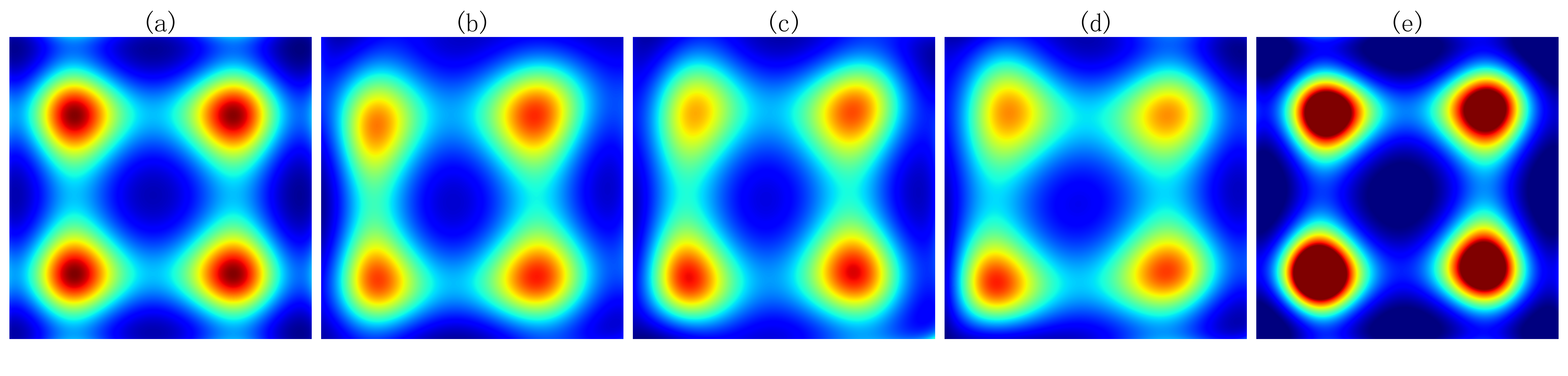}
    \caption{(Experiment 2) Inverse images from noise-free $\boldsymbol{y}$ by SBD-LAS. Images (a) and (e) are true coefficient function and inversion result based on pCN with the coarse solver,  respectively. Images (b), (c) and (d) are inversion results obtained by SBD-LAS with step ratios $\alpha=0, 0.1, 0.5$, respectively.}
    \label{fig:Experment2_estimate_SBD-LAS}
\end{figure}

\begin{figure}[H]
    \centering
    \includegraphics[width=0.8\linewidth]{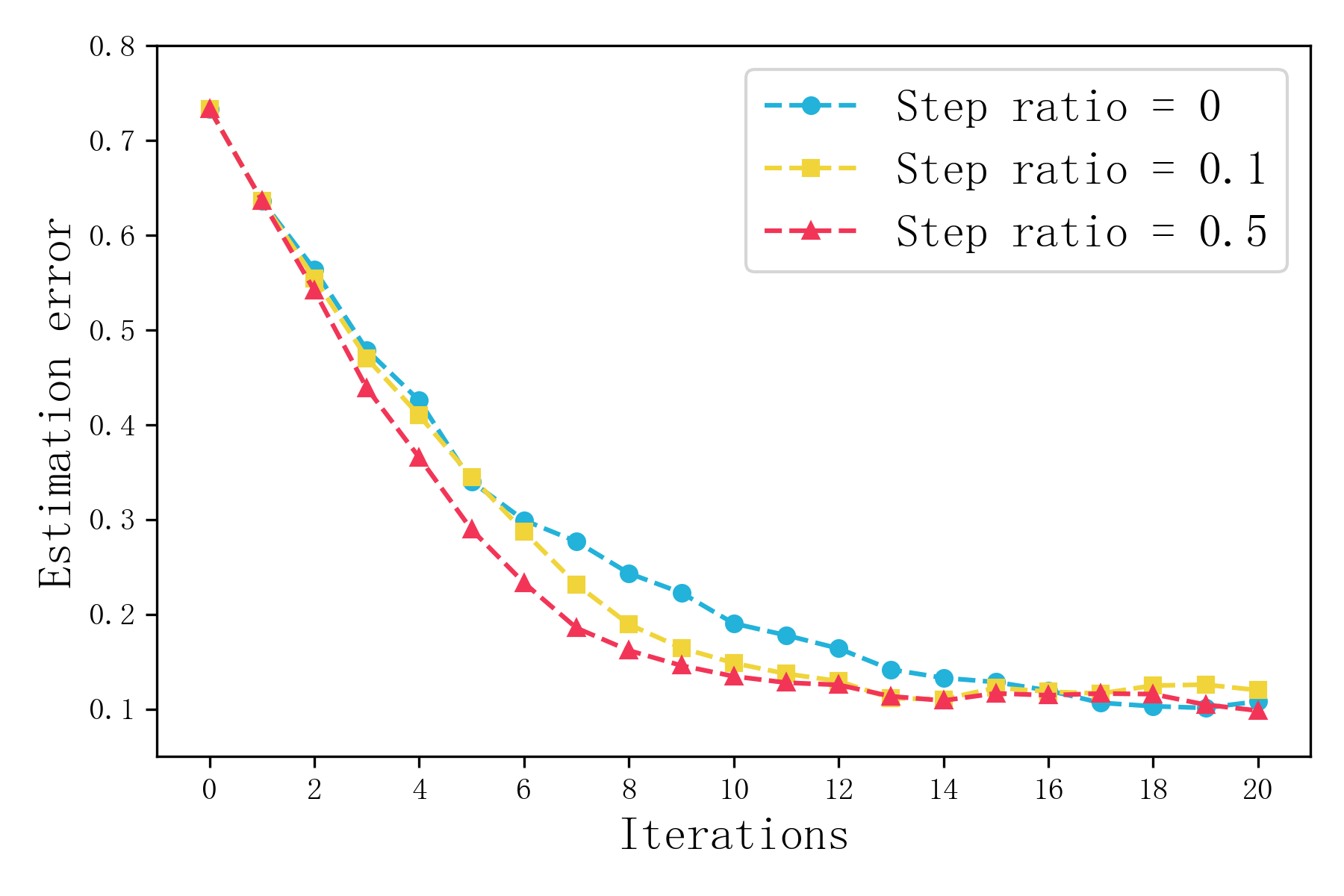}
    \caption{(Experiment 2) Errors of parameter estimation vs iterations. The three curves are from proposed method with three different step ratios $\alpha=0,0.1,0.5$. The calculation of errors follows formula (\ref{eq:MSE}).}
    \label{fig:Experment2_MSE_vs_iter}
\end{figure}

Figure \ref{fig:Experment2_estimate_SBD-LAS} presents the estimation results obtained using the SBD-LAS method. 
Similar to Experiment \ref{sec:the first experiment}, the proposed method successfully inverts the coefficient function. 
Notably, when dealing with functions exhibiting numerous peaks and valleys, the SBD-LAS method demonstrates high accuracy in capturing these features, while also requiring fewer calls to the fine solver than the direct application of fine solver in the pCN method. 
The trend in the error, as shown in Figure \ref{fig:Experment2_MSE_vs_iter}, clearly illustrates that the \textit{one-step ahead prior} significantly accelerates convergence and achieves a final accuracy comparable to that obtained with a step ratio of $\alpha=0$, while also reducing the number of calls to the fine solver. 
The figure further shows that when the step ratios are $\alpha=0.1$ or $0.5$, the algorithm converges after 12 iterations. 
In contrast, when $\alpha=0$ (i.e., without the one-step ahead strategy), convergence is only observed after the 18th iteration.
Thus, we select the results from the 14th iteration to represent the performance of the algorithm when $\alpha=0.1$ or $0.5$, and the results from the final iteration to represent the performance when $\alpha=0$. 
This demonstrates that the introduction of the \textit{one-step ahead prior} effectively accelerates the convergence process of the SBD-LAS method and reduces the computational cost associated with the fine solver. 
The parameter estimation errors and the number of fine solver calls for each method are detailed in Table \ref{tab:Experiment2_MSE_FineSolverNums}.

\begin{table}[ht]
    \centering
    \small
    \begin{tabularx}{\textwidth}{p{1cm}>{\centering\arraybackslash}p{2cm}>{\centering\arraybackslash}p{2cm}>{\centering\arraybackslash}p{2cm}>{\centering\arraybackslash}p{2cm}>{\centering\arraybackslash}p{2cm}}
    \toprule
     & \textbf{Fine solver} & \textbf{Coarse solver} & \textbf{SBD-LAS $\alpha=0$} & \textbf{SBD-LAS $\alpha=0.1$} & \textbf{SBD-LAS $\alpha=0.5$} \\
    \midrule
    $\textbf{error}$ & 0.0702 & 0.3580 & 0.1082 & 0.1104 & 0.1091 \\
    $\boldsymbol{N_f}$ & 5e5 & 0 & 2.1e4 & 1.5e4 & 1.5e4 \\
    \bottomrule
    \end{tabularx}
    \caption{(Experiment 2) The errors of parameter estimation and the numbers of calls to the fine solver $N_f$, based on different methods.}
    \label{tab:Experiment2_MSE_FineSolverNums}
\end{table}

\begin{figure}[H]
    \centering
    \includegraphics[width=0.6\linewidth]{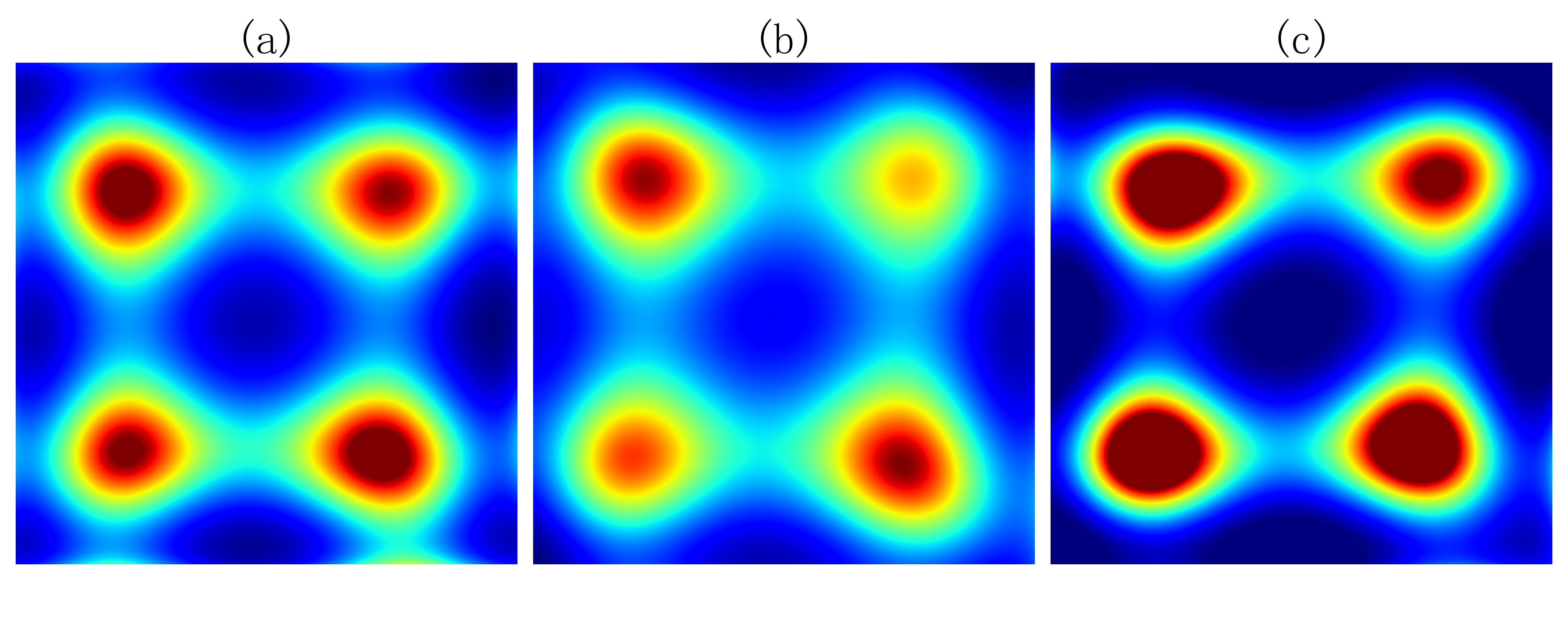}
    \caption{(Experiment 2) Inverse images of noisy observation ($\delta=0.001$). Images (a), (b), (c) are inversion results obtained by pCN with the fine solver, SBD-LAS with step ratio $\alpha=0$ and pCN with the coarse solver, respectively.}
    \label{fig:Experment2_estimates_noisy}
\end{figure}

Finally, we applied the proposed SBD-LAS method to invert noisy observational data, with a noise level $\delta=0.001$ and a step ratio $\alpha=0$ for SBD-LAS. 
The iterative process was carried out for iteration number $K=20$, with 1000 training points in each iteration, and the settings for the surrogate model were consistent with those used in the noise-free experiment. 
The estimated coefficient functions are shown in the Figure \ref{fig:Experment2_estimates_noisy} and errors of estimates for the fine solver, SBD-LAS, and the coarse solver were 0.0901, 0.0880, and 0.4279, respectively. 
These results demonstrate that our method remains effective in handling noisy observation.

\subsection{Interface Problem}
\label{sec:Interface problem}
Finally, we conducted an experiment focused on a more complex scenario: the interface problem. 
Interface problems are common in engineering, referring to boundary value problems that involve regions with different physical media or material properties. 
These interfaces typically represent boundaries between distinct materials or states, such as the contact surface between fluid and solid phases, the interface between fluids of differing densities or temperatures, and the boundaries between media with different electrical conductivity or thermal conductivity.

\begin{figure}
    \centering
    \includegraphics[width=1\linewidth]{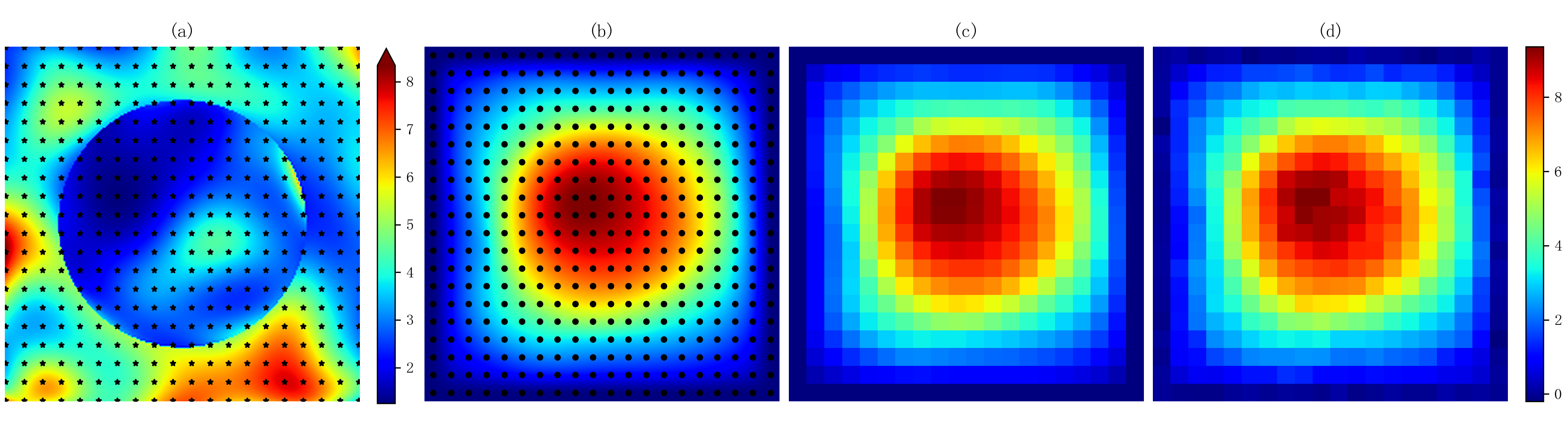}
    \caption{The true images of the third experiment: (a) is true coefficient function $a$ with black stars representing positions of parameter $\boldsymbol{\theta}$, (b) is corresponding solution $u$ with black dots representing locations of observation, (c) is noise-free observation $\boldsymbol{y}$ and (d) is noisy observation $\boldsymbol{y}^{*}$ ($\delta=0.1$).}
    \label{fig:Experment3_ture_a_u}
\end{figure}

Figure \ref{fig:Experment3_ture_a_u} displays the true coefficient function $a$, the corresponding PDE solution $u$ with the domain of definition $ \Omega = [-1,1]^2 $, noise-free observation $\boldsymbol{y}$ and noisy observation $\boldsymbol{y}^{*}$ (noise standard deviation $\delta=0.1$). 
The interface of this problem is a circular boundary defined as $S = \{(x_1, x_2) : x_1^2 + x_2^2 = 0.7^2\} $. Due to the expansion of the domain $\Omega$, the dimension of $\boldsymbol{\theta}$ for Gaussian Process Regression (GPR) has also increased, totaling 400 dimension corresponding to a 20x20 uniform grid over the domain. 
We divide elements of $\boldsymbol{\theta}$ into two sets, $\Theta_{\text{in}} = \{\theta : \theta = a(x_1, x_2), x_1^2 + x_2^2 <= 0.49\}$ and $ \Theta_{\text{out}} = \{\theta : \theta = a(x_1, x_2), x_1^2 + x_2^2 > 0.49\}$, where $(x_1, x_2)$ is the coordinate of the grid node.
$\Theta_{\text{in}}$ and $\Theta_{\text{out}}$ are used to parameterization by GPR within and outside the interface, respectively.
In this experiment, the coefficient functions both inside and outside the interface follow a log-Gaussian random field (see Equation \ref{eq:log-gaussian random field}). 
Unlike numerical experiments \ref{sec:the first experiment} and \ref{sec:second experiment}, the parameter $l$ in equation \ref{eq:log-gaussian random field} is 1 in here. 
The prior parameters $\pi_{\text{pri}}^{\text{in}}(\boldsymbol{\theta})$ and $\pi_{\text{pri}}^{\text{out}}(\boldsymbol{\theta})$ for the coefficient functions inside and outside the interface can also be derived from Equation \ref{eq:log-gaussian random field}. 
The coefficient function to be inverted is a combination of a sample from prior $\pi_{\text{pri}}^{\text{in}}(\boldsymbol{\theta})$ and a sample from prior $\pi_{\text{pri}}^{\text{out}}(\boldsymbol{\theta})$.

The experimental setup will be briefly introduced here.
The residual network used has 3 hidden layers with 500 units each and the activation function uses Sigmoid. 
The total iterations of SBD-LAS is set to $K = 10$, with $N = 50000$ pCN steps and a step size of $\beta = 0.008$, and the number of training points is $M = 1000$ for each iteration.
The coarse and fine solvers are finite element methods with 92 and 654 mesh points, respectively. 
And the initial prior $\pi_{\text{pri}}^{0}(\boldsymbol{\theta})$ is obtained by running  200,000 steps pCN with coarse solver. 

\begin{figure}[H]
    \centering
    \includegraphics[width=1\linewidth]{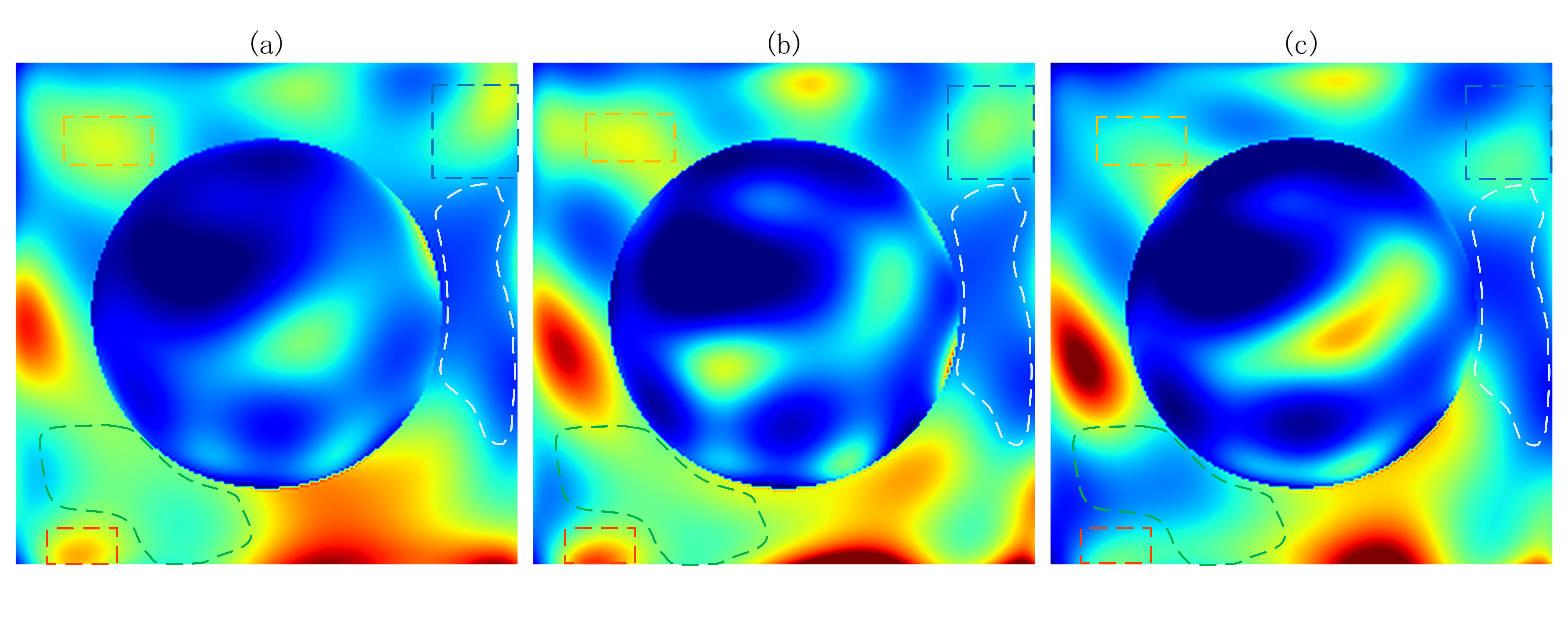}
    \caption{(Experiment 3) Inverse images of noisy observation $\boldsymbol{y}^{*}$ ($\delta=0.1$). Inverse images (a), (b) and (c) are based on pCN with the fine solver, the SBD-LAS with $\alpha=0$ and pCN with the coarse solver, respectively.}
    \label{fig:Experment3_estimate_fine_coarse}
\end{figure}

The inversion results of noisy observation $\boldsymbol{y}^{*}$ using pCN with the fine and coarse solvers are depicted in Figure \ref{fig:Experment3_estimate_fine_coarse}, with an error of 0.0742 for the fine solver and an error of 0.138 for the coarse solver. 
Next, we employ the proposed method SBD-LAS with $\alpha=0$ to invert the coefficient function $a$, with the results shown in picture (b) of Figure \ref{fig:Experment3_estimate_fine_coarse} and the error of parameter estimation is 0.116. 
As can be seen from Figure \ref{fig:Experment3_estimate_fine_coarse}, the inversion results using pCN with the fine solver have essentially learned the shape of $a$ outside the interface, specially in these dotted box area. 
Within the rectangular dashed-box region, SBD-LAS successfully captured the peaks of the coefficient function, and within the curved box region, the method accurately reconstructed the shape characteristics of the coefficient function. 
In contrast, the results from the coarse solver exhibit poorer inversion performance, failing to accurately capture the shapes outside the interface and poorly learning the function values at certain locations.
Overall, the method presented here is capable of achieving basic inversion effects, and compared to the pure coarse solver, it can better learn the shape of $a$ without overestimating the peaks.  

\section{Conclusion}
\label{sec:Conclusion}
The infinite-dimensional Bayesian inverse problem governed by PDEs is typically parameterized as a finite high-dimensional problem, which can be solved using various numerical methods. 
In high-dimensional settings, most numerical methods, such as MCMC and variational inference, involve numerous PDE solving, necessitating the use of computationally cheaper surrogate models. 
However, constructing a globally accurate surrogate model capable of accurately mapping high-dimensional complex problems demands high model capacity and large amounts of data. 
We proposes an efficient locally accurate surrogate model for Bayesian inverse problems, along with an adaptive experimental design approach, SBD-LAS, for obtaining this model. 
Numerical experiments confirm the advantages of our method in terms of both inversion accuracy and computational efficiency.

Additionally, there are also some limitations in our method. First, selecting an appropriate step ratio $\alpha$ is crucial for accelerating the SBD-LAS method. An optimal step ratio $\alpha$ can achieve the best performance in terms of acceleration. Second, given that our SBD-LAS is an iterative process, the computational cost of SBD-LAS will increase accordingly when the parameterization of the coefficient function becomes computationally expensive.

\section{Declaration of competing interest}
The authors declare that they have no known competing financial interests or personal relationships that could have appeared to influence the work reported in this paper.

\section{Acknowledgments}
The author Jinyong Ying was supported by the Natural Science Foundation of Hunan Province (Grant No. 2023JJ30648).
This work was supported in part by the High Performance Computing Center of Central South University.

\appendix
\section{Appendix 1}
\label{Appendix 1}
The following example of a one-dimensional constant coefficient ODE demonstrates the effectiveness of the locally accurate surrogate model.
\begin{equation*}
    \begin{cases}
    u^{\prime}(x)-\theta\cdot\sqrt[2]{u(x)}\cdot\cos(x)=0,&x\in(0,\pi),\\u(x)=0,&x\in\{0,\pi\},
    \end{cases}
\end{equation*}
where $u\in C([0,\pi])$, $\theta>0$ is the coefficient of the ODE to be inverted. In this example, the observed position of $u$ is 20 isometric points on $[0,\pi]$ and $\boldsymbol{y}$ is the observation vector, from which we have \textit{parameters to observations} map $\mathcal{G}\colon(0,+\infty)\to\mathbb{R}^{20}$. we can get the likelihood function
\begin{equation*}
    l(\theta \mid \boldsymbol{y}) \propto \exp\left\{-\frac12\|\boldsymbol{y}-\mathcal{G}(\theta)\|^2\right\},
\end{equation*}
where $\left\|\cdot\right\|$ is the Euclidean $L^{2}$-norm.

Let $\theta^* = 5$ to obtain an observation $\boldsymbol{y}=\{u_{1,}u_2\cdots u_{20}\}$ of $u$ at the observation positions, then the inverse problem, in this case, is inferring $\theta^*$ from $\boldsymbol{y}$. 
And we use fully connected neural network $NN{:}\mathbb{R}\to\mathbb{R}^{20}$ ($\tilde{G}=NN$) to approximate $\mathcal{G}$, which derives a surrogate of likelihood function 
\begin{equation*}
    \tilde{l}(\theta \mid \boldsymbol{y}) \propto \exp\left\{-\frac12\|\boldsymbol{y}-\tilde{G}(\theta)\|^2\right\},
\end{equation*}
where NN has 3 hidden layers with 20 units each and the activation function uses PReLU.  

Next, we train the neural network (NN) using training data of different distributions and quantities to demonstrate how the surrogate likelihood function varies with changes in the NN (training data). 
As illustrated in the example of Figure \ref{show example}, supposing the parameter space is (0,10], and the NN locally accurate model is trained with 10 data points concentrated in the high-probability region. 
The MSE of the likelihood surrogate in the high-probability region is 0.0002, which is computed by
\begin{equation}
    \label{eq:MSE_of_likelihood}
    MSE = \frac{1}{20} \sum_{i=1}^{20} \left( l(\theta_i \mid \boldsymbol{y}) - \tilde{l}(\theta_i \mid \boldsymbol{y}) \right)^2,
\end{equation}
where $\{\theta_{i}\}_{i=1}^{20}$ are 20 points evenly spaced within the interval $[4.5, 5.5]$. 
In contrast, the NN globally coarse model is trained with 10 data points distributed across the entire space, resulting in an MSE of 0.3181 in the high-probability region. 
Finally, the NN globally accurate model is trained with 100 data points spread throughout the entire space, yielding an MSE of 0.0003 in the high-probability region.

From these results, we observe that, for the same number of training points, the surrogate likelihood in the high-probability region is more accurate under the locally accurate surrogate model compared to the globally coarse surrogate model. 
To achieve the same level of accuracy in the high-probability region under a globally accurate surrogate model, several times more training samples are required. 
Additionally, it is important to note that even if the NN performs poorly in low-probability regions, the surrogate likelihood function still assigns low probabilities to these regions. 
Additionally, as shown in Table \ref{tab:example_table}, the computational cost of the globally accurate surrogate model increases as the parameter space expands, whereas the cost of the locally accurate surrogate model remains unchanged.

\section{Appendix 2}
\label{Appendix 2}
In this experiment, the system function is given by:
\begin{equation}
\label{eq:show example2}
    \begin{cases}
    y_1 = \theta_{1} + \theta_{2} + \theta_{1} \cdot \theta_{2},\\
    y_2 = \theta_{1} + \theta_{2} - \theta_{1} \cdot \theta_{2},
    \end{cases}
\end{equation}
where $\boldsymbol{y} = (y_1, y_2)$ is the system output and $\boldsymbol{\theta} = (\theta_1, \theta_2)$ is the system parameter. We preset the target parameter as $\boldsymbol{\theta}^* = (2.5, 2.5)$, and the corresponding observation $\boldsymbol{y} = (11.25, -1.25)$ is obtained by solving the system function. For simplicity, the noise-free $\boldsymbol{y}$ is directly treated as the observed data. Thus, the inverse problem in this experiment involves recovering the target parameter $\boldsymbol{\theta}^*$ from the given $\boldsymbol{y}$.

We employ a fully connected neural network as the surrogate model. The network consists of an input layer with 2 units, an output layer with 2 units, and 2 hidden layers, each comprising 40 units. The activation function used in the hidden layers is the Sigmoid function. 

We apply the method proposed in Section \ref{sec:Design point inference} to solve this inverse problem, with the results presented in Figures \ref{fig:show example2-1} and \ref{fig:show example2-2}. Figure \ref{fig:show example2-1} illustrates the trajectory of the posterior mean during the algorithm's execution, demonstrating that our method enables the posterior to progressively converge to the true posterior. Furthermore, the comparison of trajectories in Figures \ref{fig:show example2-1} and \ref{fig:show example2-2}, along with the posterior mean error curve shown in Figure 2, confirms the validity of the \textit{one-step ahead prior} and highlights its significant acceleration effect.

\bibliographystyle{unsrt}  
\bibliography{main}

\end{document}